%% file: main.tex
\documentclass[10pt,conference]{IEEEtran}

\usepackage{flushend}
\usepackage{graphicx}
\usepackage{balance}
\usepackage{comment}
\usepackage{cite}
\usepackage[active]{srcltx}
\graphicspath{{./fig/}}
\usepackage{amsmath}
\usepackage{amsfonts}
\usepackage{mathtools}
\usepackage{algorithmic}
\usepackage{float}
\usepackage{caption}
\usepackage{subcaption}
\usepackage{color}
\newcommand\norm[1]{\lVert#1\rVert}

\newcommand{\Gaussian}[0]{\ensuremath{\mathcal{N}}}

\DeclareMathOperator*{\argmax}{arg\,max}
\newcommand{\eat}[1]{}
\usepackage{soul}

\renewcommand{\baselinestretch}{0.967}

\title{Fusion of Radio and Camera Sensor Data for Accurate Indoor Positioning}

\author{
Savvas Papaioannou, Hongkai Wen, Andrew Markham and Niki Trigoni 
\vspace{1.6mm}\\
\fontsize{10}{10}\selectfont\itshape
Department of Computer Science, University of Oxford, Oxford, OX1 3QD, UK\\
\fontsize{9}{9}\selectfont\ttfamily\upshape
\{firstname.lastname\}@cs.ox.ac.uk\\
}

\begin{document}
\maketitle

\input{abstract}
\input{intro}

\input{problem}
\input{arch}
\input{algorithm}
\input{evaluation}
%\input{discussion}
\input{background}
\input{conclusion}

\section{Acknowledgment}
We would like to thank Laing O'Rourke for funding this research and also the Pitt Rivers Museum for allowing us to conduct our experiments in the museum's space.
\begin{small}
\bibliographystyle{IEEEtran}
\bibliography{main}    
\end{small}
\end{document}

%% file: abstract.tex
%%%%%%%%%%%%%%%%%%%%%%%%%%%%%%%%%%%%%%%%%%
% Abstract
\begin{abstract}
Indoor positioning systems have received a lot of attention recently due to their importance for many location-based services, e.g. indoor navigation and smart buildings. Lightweight solutions based on WiFi and inertial sensing have gained popularity, but are not fit for demanding applications, such as expert museum guides and industrial settings, which typically require sub-meter location information. In this paper, we propose a novel positioning system, RAVEL (Radio And Vision Enhanced Localization), which fuses anonymous visual detections captured by widely available camera infrastructure, with radio readings (e.g. WiFi radio data). Although visual trackers can provide excellent positioning accuracy, they are plagued by issues such as occlusions and people entering/exiting the scene, preventing their use as a robust tracking solution. By incorporating radio measurements, visually ambiguous or missing data can be resolved through multi-hypothesis tracking. We evaluate our system in a complex museum environment with dim lighting and multiple people moving around in a space cluttered with exhibit stands. Our experiments show that although the WiFi measurements are not by themselves sufficiently accurate, when they are fused with camera data, they become a catalyst for pulling together ambiguous, fragmented, and anonymous visual tracklets into accurate and continuous paths, yielding typical errors below  1~meter. %In addition, no site-specific training is required as the radio propagation model is dynamically learned. The proposed system is particularly suitable for busy indoor environments: it incurs negligible cost and deployment effort, requires no map nor training, and yet it offers accuracy akin to expensive infrastructure-based systems (e.g. UWB).
\end{abstract}

%%% Local Variables: 
%%% mode: latex
%%% TeX-master: "main"
%%% End: 

%% file: intro.tex
%%%%%%%%%%%%%%%%%%%%%%%%%%%%%%%%%%%%%%%%%%
% Introduction

\section{Introduction} 
\label{sec:intro}
In the last five years we have witnessed an unprecedented interest in indoor positioning technologies, with a variety of solutions developed in academic and industrial research labs. Although the field has reached a significant level of maturity there is still no dominant solution. As a consequence, positioning services are still lacking in many buildings. For a solution to be widely implemented and adopted it must satisfy two key requirements: low cost and high accuracy. Current low cost systems exploit or reuse existing infrastructure (e.g. WiFi) or can even operate without infrastructure (e.g. inertial tracking). These inexpensive, ubiquitous positioning approaches however cannot provide the sub-metre accuracy that high-end, expensive solutions like UWB or ultrasound can. The dichotomy between cost and accuracy has fragmented the technology landscape, leading to a plethora of competing solutions that cannot satisfy both requirements simultaneously. We present a fresh approach that seeks to unify the two disparate camps, providing high positioning accuracy with very low cost.

Our fundamental observation is that many applications that require high positioning accuracy are in large public or commercial spaces, such as airports, shopping centres, museums and industrial plants. These large public spaces are typically extensively covered with CCTV cameras for reasons of safety and security. In this paper, we propose the use of existing camera infrastructure for the originally unintended task of indoor positioning. However, although camera based tracking can provide excellent position information in ideal conditions, the challenges provided by a real deployment are numerous. In particular, most security cameras are installed to provide a large field of view, typically resulting in a bird's eye view of the scene. This top-down perspective makes it difficult to distinguish facial features and accurate identification is made even more challenging when the room is not well lit (e.g. in museums) or when people wear similar uniforms or helmets (e.g. in industrial plants). Furthermore maintaining tracking as people move behind obstacles, exit the field of view, or cross paths is an exceedingly difficult task. 

Instead of trying to overcome the limitations of camera-based tracking through increasingly sophisticated tracking algorithms, we exploit opportunistic and ubiquitous radio signals (e.g. WiFi/Bluetooth Low Energy). Although, radio-based tracking is limited in terms of accuracy, it can be used to add context to trajectories obtained from camera based tracking. At a coarse level, this provides an identifier, which can be used instead of advanced video processing techniques like face recognition. At a finer level, the sequence of radio signal strengths, albeit noisy, can be used to disambiguate between multiple possible trajectories, aiding to merge and split discontinuous traces.

Motivated by the noisy nature of both visual and non-visual sensor modalities, we explore whether we can effectively combine them to overcome each other's weaknesses. We present a generic vision+radio tracking framework, RAVEL (RAdio and Vision Enhanced Localization) that can be used both in receiver-centric (i.e. people carrying smartphones) and transmitter-centric (i.e. low-cost tags in warehouse, industrial or construction sites) applications. 

% More specifically, we focus on the following questions: Can we make use of existing camera infrastructure to provide people with an accurate positioning system without compromising their privacy? Can personal electronic data (e.g. WiFi data from a person's smartphone) be used to resolve tracking ambiguities in visual data? How should we fuse anonymous visual and personal radio data to make the most of the two modalities? What are the benefits of combining these modalities in practical settings?

So far, there have been very few attempts at combining infrastructure cameras with radio for positioning. The most relevant work, EVLoc~\cite{Zhang:MobiHoc:2012}, which also fuses radio and camera sensor data, tackles a different problem, that of matching $N$ wireless devices with $N$ visual detections, rather than inferring the location of a device, using only that device's radio data and anonymous camera data from the scene. Unlike our work, they assume a known, calibrated radio model, and have performed tests in rather ideal conditions. A complementary approach is to fuse inertial measurements with visual trajectories~\cite{Jung:INFOCOM:2010,Teixeira:PETRA:2009}, but we argue that radio measurements are the fundamental primitive that is common to many applications ranging from smartphone positioning to tracking low power tags in a warehouse; inertial data is more informative, but less ubiquitous and requires application specific signal processing to yield usable motion traces. 

To the best of our knowledge, this is the first paper that proposes a practical solution of radio-aided visual tracking, and tests it in a truly complex and realistic scenario. Specifically, our contributions are:

\begin{itemize}
 \item We provide a fresh perspective on the problem of low-cost high-accuracy positioning in large open-plan indoor spaces, enabled by the fusion of anonymous visual data and radio data. 
 %\item We formulate the problem of tracking people by combining visual- and radio-based positioning modalities. We highlight the key challenges that we have encountered in a complex indoor museum environment, and explain why existing approaches are not designed to cope with reality.
 \item We design a novel multi-hypothesis probabilistic approach (RAVEL) to fuse radio and camera data that is robust to noisy and incomplete measurements.
 \item We show how the radio propagation model for a particular environment can be learnt online, requiring no site-specific surveying.
 \item We evaluate the proposed approach in a museum setting, and compare it with multi-hypothesis tracking which only uses visual data.\\
\end{itemize}

The remainder of this paper is structured as follows: Sec.~\ref{sec:problem} outlines the requirements of a system for fusing visual and radio data to accurately track users in indoor environments. Sec.~\ref{sec:arch} presents the overview of our radio-aided visual tracking system, while Sec.~\ref{sec:algorithm} discusses the details of the proposed algorithms. Sec.~\ref{sec:evaluation} evaluates the proposed system in a real-world museum scenario. Sec.~\ref{sec:background} overviews related work and Sec.~\ref{sec:conclusion} concludes the paper and discusses future work.

%% file: problem.tex
\section{System Requirements}
\label{sec:problem}

%\subsection{Requirements}
%\label{ssec:problem-requirements}
Here we lay out the key requirements and challenges for a practical visual- and radio-based positioning solution.
% in complex indoor settings, which have not been fully addressed by the existing approaches, and drive the design of our system.

\noindent \textbf{Lightweight: } 
The proposed solution should be lightweight enough to run in real-time on resource-constrained devices with minimum training effort. This excludes computationally-expensive approaches, such as part-based object detectors (e.g.~\cite{Felzenszwalb:PAMI:2010}), or face recognition techniques.

\noindent \textbf{Accurate: } 
We aim for \emph{sub-meter} tracking accuracy for a particular user, as required by many applications. Existing radio-based positioning approaches typically cannot achieve such accuracy, while visual tracking techniques tend to generate \emph{anonymous} and often segmented trajectories that cannot be assigned to a particular user.

\noindent \textbf{Robust: }
The proposed solution should be usable in busy and dynamic indoor environments, and robust to noisy visual / radio data. In practice, fixed obstacles such as furniture may easily block line of sight to both the camera and the radio access points, resulting in occlusions and attenuation respectively. Changes in illumination and arbitrary movements of people can also have catastrophic effects on approaches relying on perfect visual detections.

%%% Local Variables: 
%%% mode: latex
%%% TeX-master: "main"
%%% End: 

%% file: arch.tex
%%%%%%%%%%%%%%%%%%%%%%%%%%%%%%%%%%%%%%%%%%
% System arch

\begin{figure}
\includegraphics[width=\columnwidth]{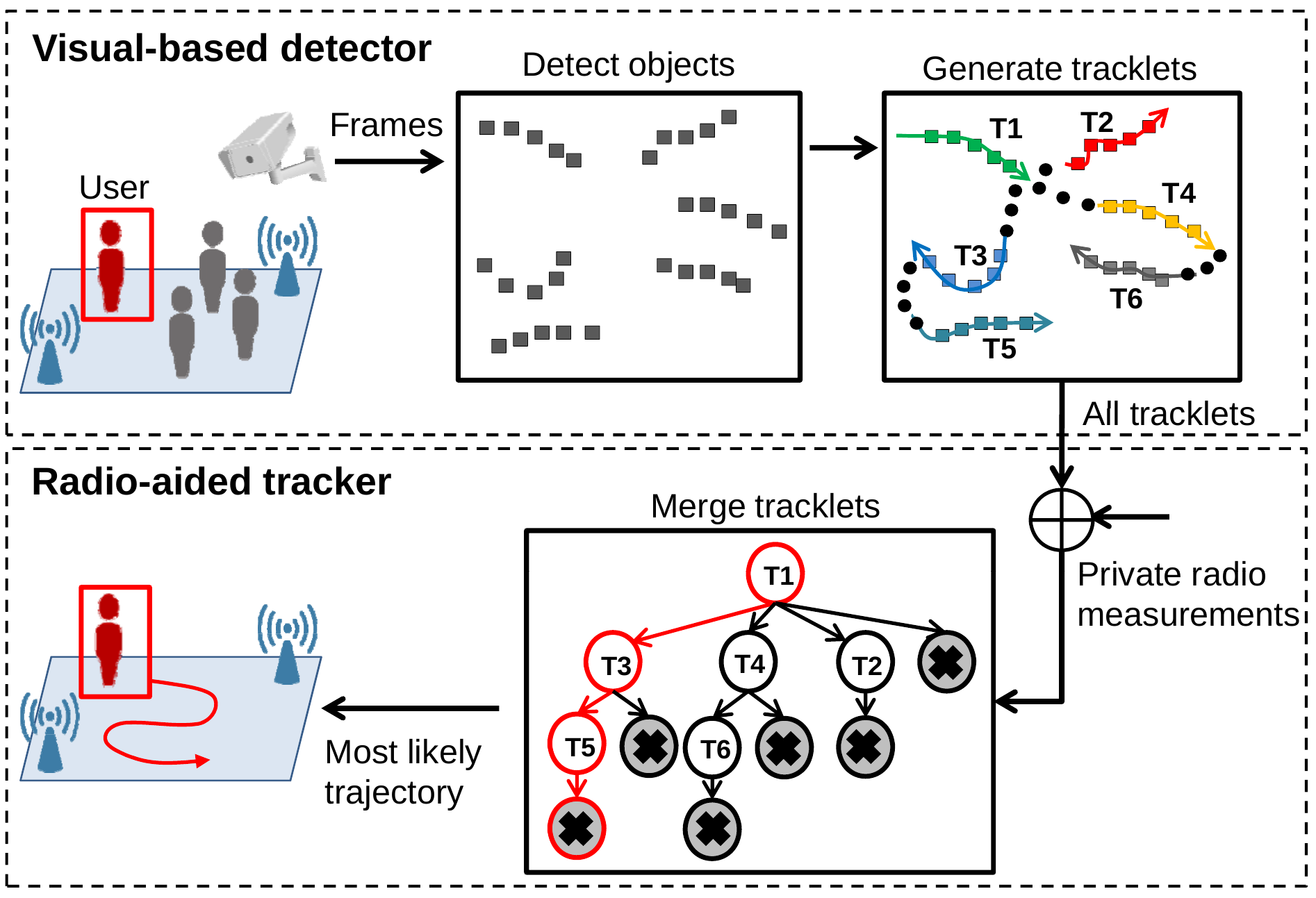}
\caption{Overview of the proposed RAVEL system.}
\label{fig:arch}
\end{figure}

\section{System Overview}
\label{sec:arch}
We are now in a position to present the proposed Radio And Vision Enhanced Localization (RAVEL) system, which consists of two components: a \emph{visual based detector} and a \emph{radio aided tracker} shown in Fig. (1). These components can run in the same device (e.g. a server collecting measurements from multiple transmitters) or across different devices (e.g. smart cameras running the visual-based detector and disseminating traces to mobile devices, each running their own radio-aided tracker to preserve user privacy). The rest of the paper considers the latter, more challenging case of distributed tracking, but the algorithms presented are equally applicable to centralized tracking.

\subsection{The Tracking Problem}
\label{ssec:problem-problem}
Let us assume the indoor environment is monitored by a calibrated (known extrinsic and intrinsic parameters) stationary camera (the proposed approach can be easily extended to multi-camera scenarios). For a given time window of size $W$, $W \in \mathbb{Z}$, the camera captures a series of frames $[f_1,...,f_W]$ within its field of view (FOV). We assume that each frame $f_i$ contains a number of camera detections of moving objects within the FOV, denoted as $C_i=\{ c^1_{i},...,c^j_{i},...\}$, $1\leq i \leq W$, $1\leq j \leq |C_i|$. A camera detection $c^j_i$ is represented as a bounding box of the detected object, or simply as the coordinates of the center of the bounding box, i.e. $c^j_i=(x^j_i,y^j_i)$.

We also assume that at each time $i$, the mobile device carried by a particular user \emph{receives} a set of radio measurements $r_i = \{ r^1_{i},...,r^m_{i},...\}$, where $r^m_i$ is the Received Signal Strength (RSS) measurement of the $m$-th radio basestation at time $i$. The problem is is to \emph{estimate the trajectory of a user given the sequence of anonymous camera detections $[c_1,...,c_W]$ and personal radio measurements $[r_1,...,r_w]$}.

\subsection{Visual-based Detector}
\label{ssec:arch-visual}
\begin{figure}
\includegraphics[width=\columnwidth]{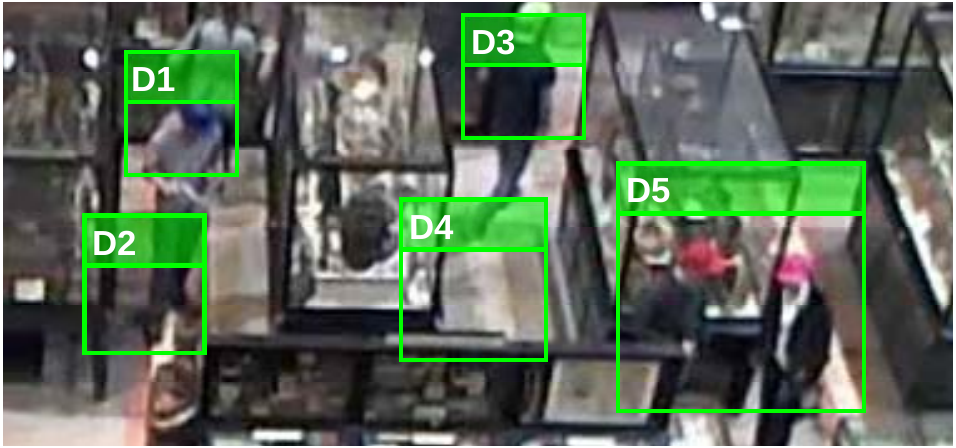}
\caption{Three cases of noisy detections generated by our visual-based detector: a) multiple detections are generated for one moving target (D1 and D2), b) a detection contains no moving targets at all (D4), and c) one detection contains multiple moving targets (D5).}
\label{fig:detections}
\end{figure}

The visual-based detector processes the captured camera footage as follows. For each frame $f_i$, a set of anonymous camera detections $C_i=\{ c^1_{i},...,c^j_{i},...\}$ of the interesting objects (i.e. moving people) are firstly extracted. In our implementation, we use a lightweight MoG-based background subtraction approach~\cite{Stauffer:CVPR:1999} to detect moving people, which does not require heavy training and with some optimization it can run in real-time in embedded camera networks~\cite{Shen:SenSys:2012}. %\eat{Grayscale footage, rather than color footage, is used in the detector for reasons of efficiency and maximum generalization (many existing CCTV systems are not color). }
However, the computed set of detections $C_i$ can be very \emph{noisy} and \emph{unreliable}. Fig.~(\ref{fig:detections}) shows a frame with three typical cases of noisy detections observed in our experiments, namely multiple detections from a single person (splitting); single detections corresponding to multiple people (merging); and null detections (empty bounding box, caused by a visual artifact such as a moving shadow). In this case, linking the detections into coherent trajectories is not a trivial task. Note that the visual-based detector only links detections into short trajectory segments (referred to as \emph{tracklets} hereafter) when they unambiguously belong to the same target, as discussed in Sec.~\ref{ssec:algorithm-generation}. The tracklets could be broadcasted to phones in a number of ways, including overloading the WiFi beacon frames (i.e. beacon stuffing)~\cite{Chandra2007}, a technique that could also be used to preserve the user's privacy.

\subsection{Radio-aided tracker}
\label{ssec:arch-radio}
The tracklets generated by the visual-based detector tend to be very short, and are still \emph{anonymous}: we have no knowledge of which tracklet should belong to a given user. Therefore, the proposed system broadcasts the generated tracklets (which are just streams of two dimensional coordinates) to the radio-aided tracker, which runs on the mobile devices carried by the users. With the received anonymous tracklets, the radio-aided tracker cross references the user's private radio footprint, i.e. radio measurements from the known basestations, to determine which tracklets should be selected and merged to produce the complete trajectory of this particular user. To achieve this, our tracker uses a multi-hypothesis tracklet merging approach. In a nutshell, it builds \emph{tracklet trees} that encode all hypotheses of user paths. We are now in a position to present the details of the proposed algorithms.

%%% Local Variables: 
%%% mode: latex
%%% TeX-master: "main"
%%% End: 

%% file: algorithm.tex
%%%%%%%%%%%%%%%%%%%%%%%%%%%%%%%%%%%%%%%%%%
% Algorithm
\section{Proposed Algorithm}
\label{sec:algorithm}
In this section, we discuss the core algorithms of the RAVEL tracking system: the \emph{tracklet generation algorithm} (Sec.~\ref{ssec:algorithm-generation} ) used in the visual-based detector, and the \emph{tracklet merging algorithm} (Sec.~\ref{ssec:algorithm-merging}), which is the key competency of our radio-aided tracker.

\subsection{Tracklet Generation Algorithm }
\label{ssec:algorithm-generation}
Given a fixed window of frames $[f_1,...,f_W]$, and extracted sets of camera detections $[C_1,...C_W]$, the task of the tracklet generation algorithm is to link the elements of $C_i$-s into unambiguous trajectory segments, referred to as tracklets. The proposed algorithm leverages a realistic model of human motion to ascertain that a sequence of detections from consecutive frames belong to the same person with high certainty, and can thus be grouped together into a tracklet $\tau$. It starts by considering detections in  the first frame and proceeds in chronological order until all detections are grouped into a set of tracklets $\mathcal{T}=\{ \tau_1,...,\tau_N\}$. 

More specifically, at the beginning, or when an existing tracklet cannot be further extended with new camera detections, a new tracklet is initiated with the first available (not-visited) camera detection, say $c_i$. To extend the tracklet with a second detection, we first check if there is a detection $c_{i+1}$ in the next frame such that $\norm{c_i-c_{i+1}} < \mathrm{DT} $ where $\mathrm{DT}$ is the maximum displacement of a target in the period between two consecutive frames. If there is no such detection, or if there are more than one such detections, $c_i$ becomes a singleton tracklet. Otherwise the two detections must belong to the same person, so they are grouped together in the same tracklet. 

Once we have at least two detections in the same tracklet, we can extend it with another detection taking into account the fact that humans normally avoid abrupt changes of direction and speed~\cite{Sethi:PAMI:1987}. For example, consider the last two detections in the tracklet (say, $c_i$ and $c_{i+1}$), and the potential to extend them with one of the available detections in the next frame, say $c_{i+2}$. In order to assess its suitability, we measure changes in the direction and in the speed of motion, and combine them to obtain a metric of confidence that the three detections concern the same person:
\begin{equation}
\label{eq:motion-model}
Q (c_{i+2}; c_i,c_{i+1}) = w_d Q_d + w_s Q_s
\end{equation}

\noindent where $Q_d$ is the cost of direction change while $Q_s$ is the cost of speed change, weighted by $w_d$ and $w_s$ respectively. In our case, $Q_d$ and $Q_s$ are defined as:
\begin{equation}
\label{eq:costs}
\begin{split}
Q_d & = 1- \frac{(c_{i+1} - c_i) \cdot ( c_{i+2} - c_{i+1} )} { \norm{c_{i+1} - c_i} \norm{c_{i+2} - c_{i+1}} } \\
Q_s & = 1 - 2 \frac{\sqrt{\norm{c_{i+1} - c_i} \norm{c_{i+2} - c_{i+1}}}}{\norm{c_{i+1} - c_i} + \norm{ c_{i+2} - c_{i+1}}}
\end{split}
\end{equation}

%From the above, we can see that $Q_d$ is actually the cosine of the angle between the displacement vectors $(c_{i+1} - c_i)$ and $(c_{i+2} - c_{i+1})$, which accounts for changes in the motion direction. On the other hand, $Q_s$ is the ratio between the geometric and arithmetic mean of the magnitude of the displacement vectors. It measures the variation in the speed of motion and penalizes those detections that are very far away from the current tracklet (note that $Q_s$ becomes 0 when the lengths of the two displacement vectors are the same). As shown above, the algorithm uses Eqn.~\eqref{eq:motion-model} to evaluate the smoothness of motion between the last two detections of the current tracklet and a candidate detection in the next frame. 

If at most one detection is found with cost $Q$ less than a predefined threshold then we include that detection in the tracklet and continue to the next frame; otherwise we stop extending this particular tracklet. The above procedure is repeated until all detections are grouped into tracklets.

\subsection{Tracklet Merging Algorithm}
\label{ssec:algorithm-merging}
Given the generated tracklets, the tracklet merging algorithm, which is the core of our radio-aided tracker, attempts to decide which tracklets should be merged together to produce the complete trajectory of the particular user. It first builds a set of tracklet trees that encode all possible trajectory hypotheses, and then searches for the trajectory hypothesis that is most consistent with the user's radio data.

\begin{figure}
\includegraphics[width=\columnwidth]{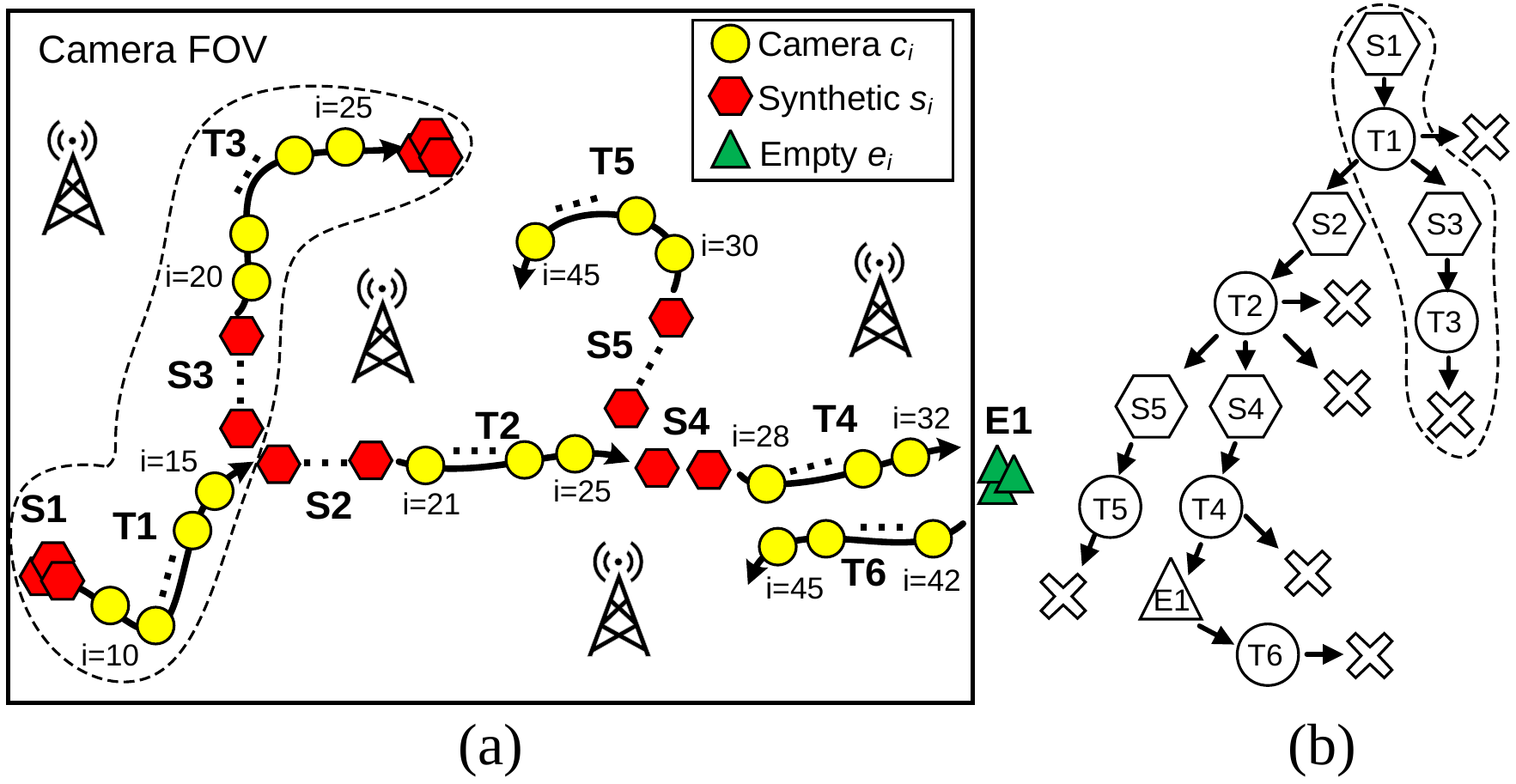}
\caption{(a) The tracklets generated by our algorithm, where the gaps are filled by synthetic and empty detections. (b) The built tracklet tree with nodes containing camera (circle nodes), synthetic (hexagon nodes), and empty (triangle nodes) detections. Each camera node is also attached with a EOT node.}
\label{fig:tracklet-tree}
\end{figure}

\noindent \textbf{Build the tracklet trees: }
Let us denote the set of tracklets generated within a window of frames $[f_1,...,f_W]$ as $\mathcal{T}=\{ \tau_1,...,\tau_N\}$. We first define $\mathcal{T}_{start}$ as the subset of tracklets in $\mathcal{T}$ that start early (before a certain time threshold) and thus could be used to start a user's trajectory. For each of those candidate tracklets, the algorithm initiates a tracklet tree. The algorithm then expands these trees as follows: for a given tree node (parent tracklet), its children become all the tracklets that start soon after it finishes and for which their first detection is at close proximity to the parent tracklet's final detection. This spatio-temporal threshold  constrains the size of the tracklet trees and can be adjusted to allow this module to run in real-time on mobile devices.

Now the tracklets in $\mathcal{T}$ are topologically connected into paths, but these paths are still incomplete. For example, consider tracklets T1 and T3 in Fig.~(\ref{fig:tracklet-tree}a), which form a path in the tree shown in Fig.~(\ref{fig:tracklet-tree}b) (ignore the nodes other than T1 and T3 for now). For this path, we can see that: a) it does not start from the beginning of time window (T1 starts at time 10); b) it ends early (T3 ends at time 25 while the window size is 45); and finally c) there is a gap of missing detection in between (the final detection in T1 is at time 15, and the first detection in T3 is at 20).

To address this, we introduce two extra types of detections, \emph{synthetic detections} $s_i \in S$ and \emph{empty detections} $e_i \in E$, in addition to the \emph{camera detections} $c_i \in C$. The synthetic detections are used to address the problem of missing camera detections when the user is actually within the camera's FOV, which are caused by occlusions or pauses of the user. On the other hand, the empty detections are introduced to account for situations where the user exits and possibly reenters the camera's FOV, and temporarily has no known coordinates. 

Concretely, our algorithm completes paths with synthetic and empty detections according to the following rules: \emph{First tracklet rule:} If the first tracklet of a path does not start at time 1 (of the window), we precede it with empty detections if it starts at the boundary of the scene, or with synthetic detections (positioned at the location of its first detection) if it does not start at the boundary. \emph{Gap rule:} If there is a gap between a parent and a child tracklet in the tree, we fill the gap with synthetic detections positioned at interpolated points between the final detection of the parent tracklet and the first detection of the child tracklet. In the unusual case that the parent tracklet ends at the boundary of the FOV and the child tracklet begins at the boundary, we fill the gap with empty (instead of synthetic) detections. \emph{Last tracklet rule:} If the last tracklet of a path does not end at the end of the window, we extend it with empty detections if it ends at the boundary of the scene, or with synthetic detections (positioned at the location of its last detection) if it does not end at the boundary. We also add an end of trajectory (EOT) node to every node containing camera detections, indicating the user may stop there.

After applying these rules, every path of the built tracklet tree represents a possible trajectory that contains $W$ consecutive detections whether camera, synthetic or empty. For example, the highlighted path in Fig.~(\ref{fig:tracklet-tree}b) corresponds to the highlighted trajectory in Fig.~(\ref{fig:tracklet-tree}a), which indicates that the user moves from the bottom left corner of the FOV to the top middle. We refer to such a trajectory as a \emph{hypothesis}, denoted as $H = [h_1,...h_W]$, where $h_i \in C \cup S \cup E$.

\noindent \textbf{Search the trackless trees: }
Once we have identified all trajectory hypotheses, we proceed to identify the most likely one. Specifically, the likelihood score of a hypothesis is defined as the sum of two parts: 
\begin{equation} 
\label{eq:likelihood-function}
L(H, R, \lambda) = L^v(H) + L^r(H, R,\lambda)
\end{equation}

\noindent where $L^v(H)$ is the \emph{visual-based likelihood score} of the hypothesis $H$, while $L^r(H, R,\lambda)$ is the \emph{radio-based likelihood score} given observed radio measurements $R$ and a radio propagation model $\lambda$. Therefore, our algorithm evaluates the likelihood score of a hypothesis in two steps: 1) it firstly estimates the user trajectory $X = [x_1,...,x_W]$ based on vision data in hypothesis $H$, and evaluates the likelihood of vision data; 2) it then evaluates the likelihood of the radio data given the estimated user trajectory $X$. The first step is already applied by existing multiple hypothesis tracking approaches (e.g.~\cite{Reid:TAC:1979,Blackman:AESM:2004}), which neglect radio data and select the hypothesis that maximizes $L^v(H)$ only. We could of course derive the overall likelihood score in one step, using a Bayesian filter to jointly fuse vision and radio data. Our design choice to separate these steps stems from our desire to reuse existing implementations of vision-only trackers and extend them flexibly with a new step that additionally exploits radio data. Now we explain how $L^v(H)$ and $L^r(H, R,\lambda)$ are evaluated in detail.

\noindent \emph{Evaluate $L^v(H)$: }
For each hypothesis $H$, our algorithm maintains a filter, to estimate the trajectory $X$ of the user. This can be implemented using a variant of a Bayesian filter, such as a Kalman filter~\cite{Kalman:KF:1960}. Let $x_i^-$ be the estimated position (state) of the user given the detections $h_{1:i-1}$, with covariance $P_i^-$. For a non-empty detection $h_i$, we define its incremental visual-based score as:
\begin{equation}
\label{eq:v-delta-likelihood}
\Delta L^v_i = 
\begin{dcases*}
  \log [p_v f(h_i; x_i^-, P_{i}^-, \theta)] &, $h_i\in C$  \\
  \log [(1-p_v) f(h_i; x_i^-, P_{i}^-, \theta)]&, $h_i\in S$\\
\end{dcases*}
\end{equation}

\noindent where $p_v$ is the constant likelihood of having a camera detection when the user is in the FOV. $f$ is a function that evaluates the likelihood of $h_i$ given the estimated $x_i^-$, $P_{i}^-$ and the parameters $\theta$ of the filter, i.e. it assesses how $h_i$ agrees with the state estimated by the filter. Then $L^v(H)$ is the normalized sum of all $\Delta L^v_i$:
\begin{equation}
\label{eq:v-likelihood}
L^v = |H|^{-1} \sum_{i=1}^{W} \Delta L^v_i
\end{equation} 

\noindent where $|H|$ is the number of non-empty detections in the hypothesis $H$.

\noindent \emph{Evaluate $L^r(H, R, \lambda)$: }
Let us assume that radio model $\lambda$ in the indoor environment can be described with the log-normal shadowing model with parameters $\{P_0, n, \sigma\}$, and the Received Signal Strength (RSS) $r(a)$ at a point which is $a$ meters to the known basestation is given by:
\begin{equation} 
\label{eq:radio-model}
r(a)=P_0 - 10 n \log_{10}(a) + \chi_{\sigma^2}
\end{equation}

\noindent where $P_0$ is the RSS measurement at a reference distance of 1 meter, $n$ is the path loss factor, and $\chi_{\sigma^2} \sim \Gaussian(0,\sigma^2)$ is the random shadowing variation. Let $x_i^+$ be the state estimated with detections $h_{1:i}$ from the previous step, which is the best guess on the location of a user at time $i$. For a non-empty detection $h_i$, we define its incremental radio-based likelihood score as:
\begin{equation}
\label{eq:r-likelihood}
\Delta L^r_i = \sum_{m=1}^{M} \log f_\Gaussian[r(dist(x_i^+, B_m)); r^m_i, \sigma^2]
\end{equation} 

\noindent where $dist(x_i^+, B_m)$ is the distance between $x_i^+$ and basestation $B_m$, and $r(dist(x_i^+, B_m))$ is the expected RSS value at $x_i^+$ (given by Eqn.~\eqref{eq:radio-model}). $r^m_i\in r_i$ is the received RSS measurement of $B_m$ on the user's mobile device. $f_\Gaussian$ is Gaussian probability density function. Therefore, the score of $h_i$ is determined by how its corresponding state estimate $x_i^+$ fits the measured $r_i$. Then the score for entire hypothesis $H$ and radio model $\lambda$ is the normalized sum of all $\Delta L^r_i$:
\begin{equation}
\label{eq:r-likelihood}
L^r = |H|^{-1} \sum_{i=1}^{W} \Delta L^r_i + \log L_\lambda
\end{equation} 

\noindent where $L_\lambda$ is the likelihood of radio model $\lambda = \{P_0, n\}$ given prior distributions on these two parameters. Our algorithm does not require perfect knowledge of $\lambda$, but attempts to learn the best model by a search through the parameter space. For simplicity, we assume the parameter $\sigma$ is known, and only vary $P_0$ and $n$. In our experiments, we assume $P_0$ and $n$ are independent, and follow the raised cosine priors governed by known hyper-parameters. Sec.~\ref{sec:evaluation} will show how in practice our algorithm finds the best radio model that is very close to the actual one.

To sum up, for each hypothesis $H$ and possible radio model $\lambda$, our algorithm evaluates the likelihood score $L(H, R, \lambda)$ as shown above, and finds the best solution $\{H^*, \lambda^*\}$ which is given by: $ \{H^*, \lambda^* \}=  \argmax_{H, \lambda} L(H,R,\lambda)$.

%\flushcolsend
%%% Local Variables: 
%%% mode: latex
%%% TeX-master: "main"
%%% End: 

%% file: evaluation.tex
%%%%%%%%%%%%%%%%%%%%%%%%%%%%%%%%%%%%%%%%%%
% Evaluation

\section{System Evaluation}
\label{sec:evaluation}

\subsection{Experimental Setup}

\begin{figure}
\centering
\includegraphics[width=\columnwidth]{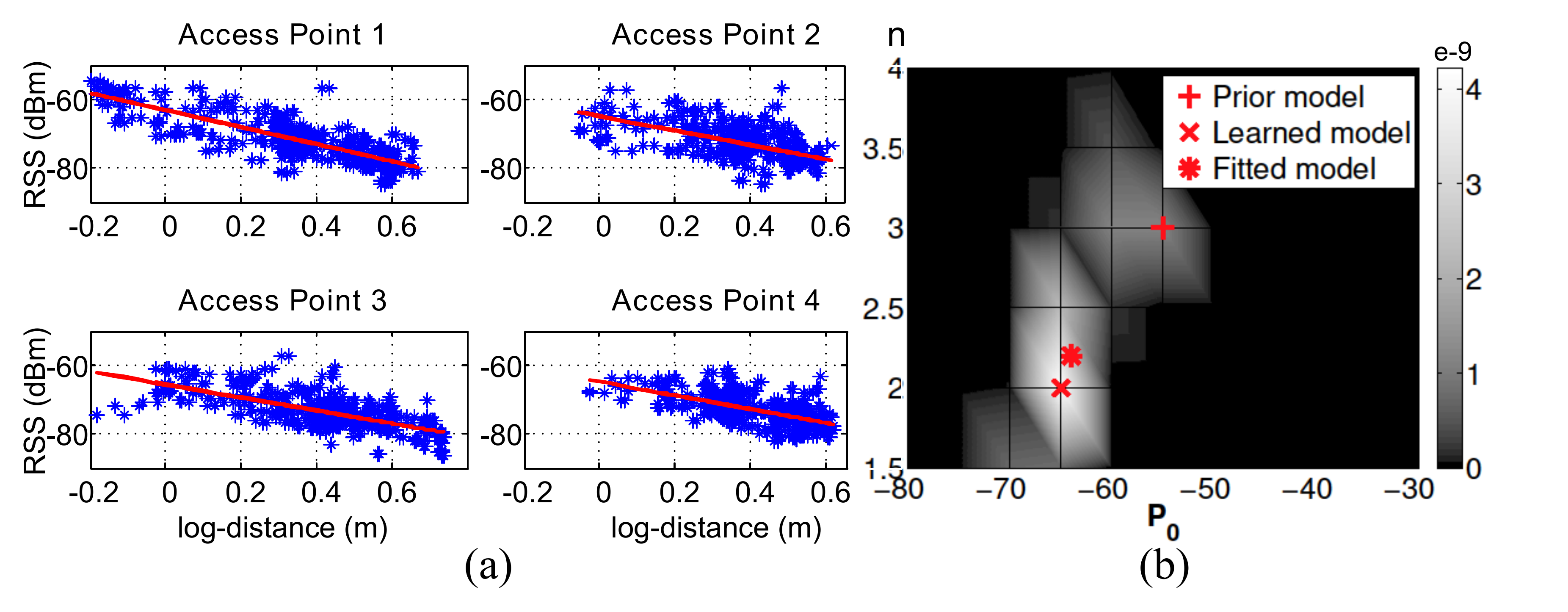}
\caption{(a) Fitted radio model from 4 APs. (b) Learning the radio model parameters by searching the parameter space.}%
\label{fig:learnModel}
\end{figure}

We have conducted a real world experiment in a three-story Museum building to evaluate the performance of the proposed approach. In our experiment we placed the camera (i.e. off-the-shelf webcam) 10m above the ground in the third floor of the building facing down covering a 11m $\times$ 12m area. All people to be tracked were walking on the first floor while looking at the museum exhibits. The duration of our experiment was 40 minutes with the camera taking video at 2fps with a resolution of 960 $\times$ 720 px. The total number of people in the scene was varying as the museum visitors were entering and leaving the scene. The minimum number of people in one frame was 8 and the maximum 20 with 4 of them having WiFi enabled smartphones. The objective of the experiment is to determine the accuracy with which people with WiFi data can track themselves in a busy environment with several other people. The WiFi measurements were taken by smartphones receiving beacons from 4 APs at a default rate of 2 samples/sec. 

In order to obtain accurate ground truth trajectories from the video footage we supplied all four people carrying WiFi smartphones with hats of different colors. Then we used a mean-shift tracker \cite{Ning2012:CVIET:2012} to track the colored hats and label the ground truth trajectories of all 4 people for our entire 4800 frame dataset. The color features were only used to acquire accurate ground truth; and both our approach and the competing approach use only gray-scale images. We noticed that in the absence of these distinctive hats, appearance features are not informative enough to tell apart one person from another, due to the dim lighting and the fact that with the downward-facing camera we did not get a view of distinctive face/body features. 

\noindent {\bf Algorithms:}  The competing approach, referred to as \emph{Vision-only tracker}, only uses the smoothness of motion to connect visual detections into trajectories; it is a multi-hypothesis tracking approach widely used by the vision community~\cite{Blackman:AESM:2004,Reid:TAC:1979}. On the other hand, our proposed approach, \emph{RAVEL} (Radio And Vision Enhanced Localization) ravels out ambiguous visual threads by exploiting both the smoothness of motion and radio signal strength data. 

\noindent {\bf Performance metrics:} We distinguish between two main operating contexts: 1) offline case, in which the competing/proposed algorithms are given data over a time window of size $W$, and are tasked to estimate the trajectory of a target during that period; and 2) online case, in which the task is to estimate the current location of the target given a window of historical data of size $W$. %For application scenarios that can tolerate some delay in location estimation, we introduce a look-ahead delay tolerance parameter, and observe the impact of that parameter on the location estimation accuracy.

For the offline case, we use two key metrics to evaluate our approach - offline location error and overlap error. \emph{Offline location error} is an accuracy metric that reflects the distance between the estimated trajectory (using the competing or proposed techniques) and the ground truth trajectory (derived by using distinctive visual markers). In the results below, we measure it as the average Euclidean distance between the ground truth and estimated detections over time. In measuring the distance between trajectories, we only use timestamps (frames) for which both the estimated and ground truth trajectories report the target to be within the field of view. 

%\begin{figure}
%\centering
%\includegraphics[width=\columnwidth]{cdfOVERLAP.pdf}
%\caption{Cumulative distribution function (CDF) of the offline and overlap errors}%
%\label{fig:cdfOfflineError}
%\end{figure}

The second offline metric, referred to as \emph{overlap error}, measures the overlap between the ground truth and estimated trajectories with respect to the camera's field of view. Let FP (false positives) be the ratio of frames in a window of size $W$ for which the estimated trajectory reports a detection in the FOV, whereas the ground truth trajectory suggests that the target is outside the FOV. Let FN (false negatives) be the ratio of frames in which the estimated trajectory reports an empty detection (outside the FOV), whereas the ground truth trajectory includes a detection within the FOV. The \emph{overlap error} between the two trajectories is defined as the sum of false positive (FP) and false negative (FN) ratios, and it reflects the percentage of time in which the estimated algorithm misclassifies the target to be in the FOV when they are not or vice versa.

For the online case, our accuracy metric is referred to as \emph{online location error}, and is the Euclidean distance between the last pair of detections of the estimated and ground truth trajectories (those at the end of the a historical window). 

%time $t$ given a historical window of information from frame $t-w$ to frame $t$, and a look ahead delay tolerance window of varying size $tol$. The results that we show in this case run the proposed and competing algorithms with data in the time window $[t-w,t+tol]$, and measure the Euclidean distance between the estimated and ground truth detections at time $t$. 

\begin{figure*}
\centering
\begin{subfigure}{0.68\columnwidth}
\includegraphics[width=\columnwidth]{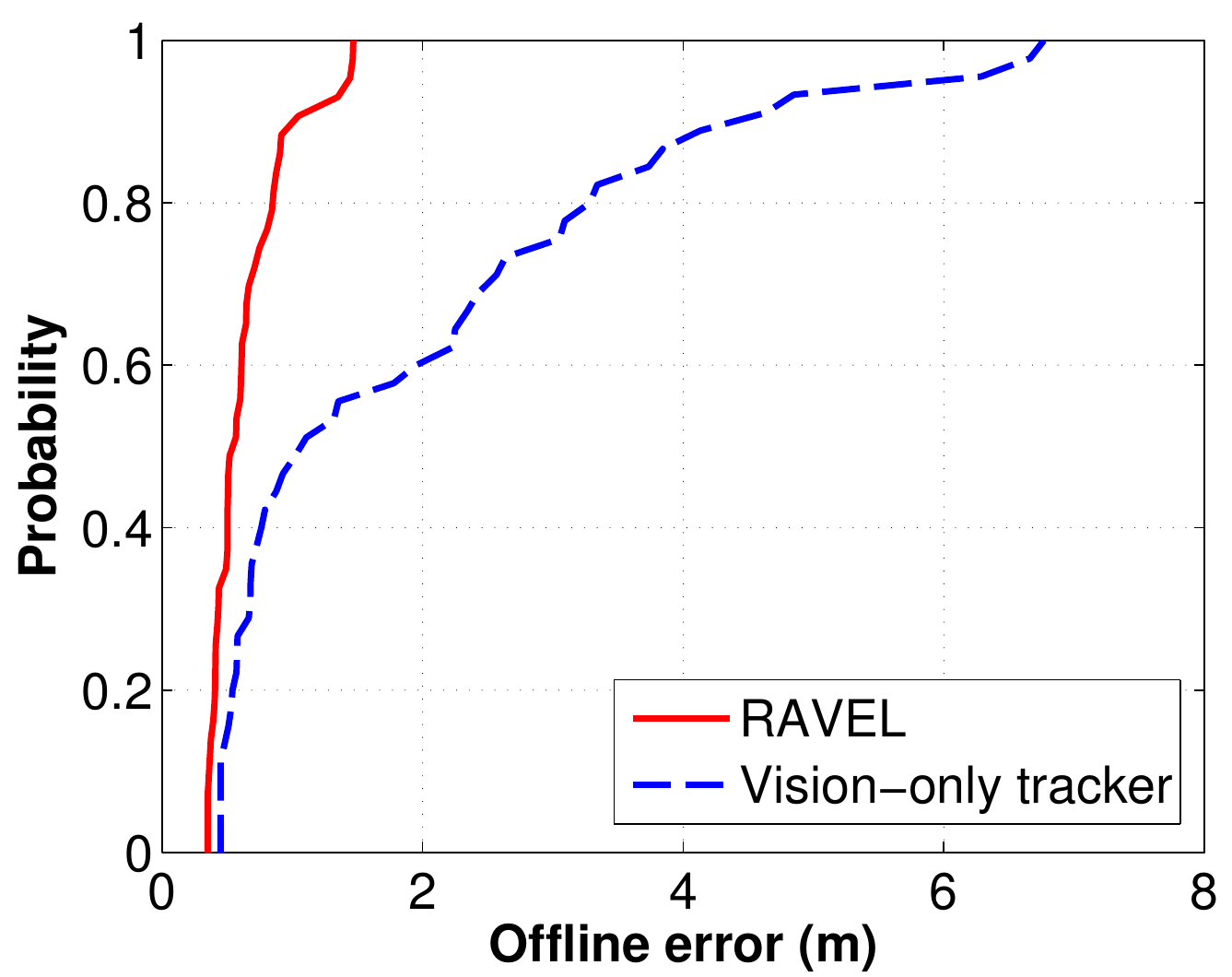}%
\caption{}%
\label{fig:cdfOfflineError}%
\end{subfigure}\hfill%
\begin{subfigure}{0.68\columnwidth}
\includegraphics[width=\columnwidth]{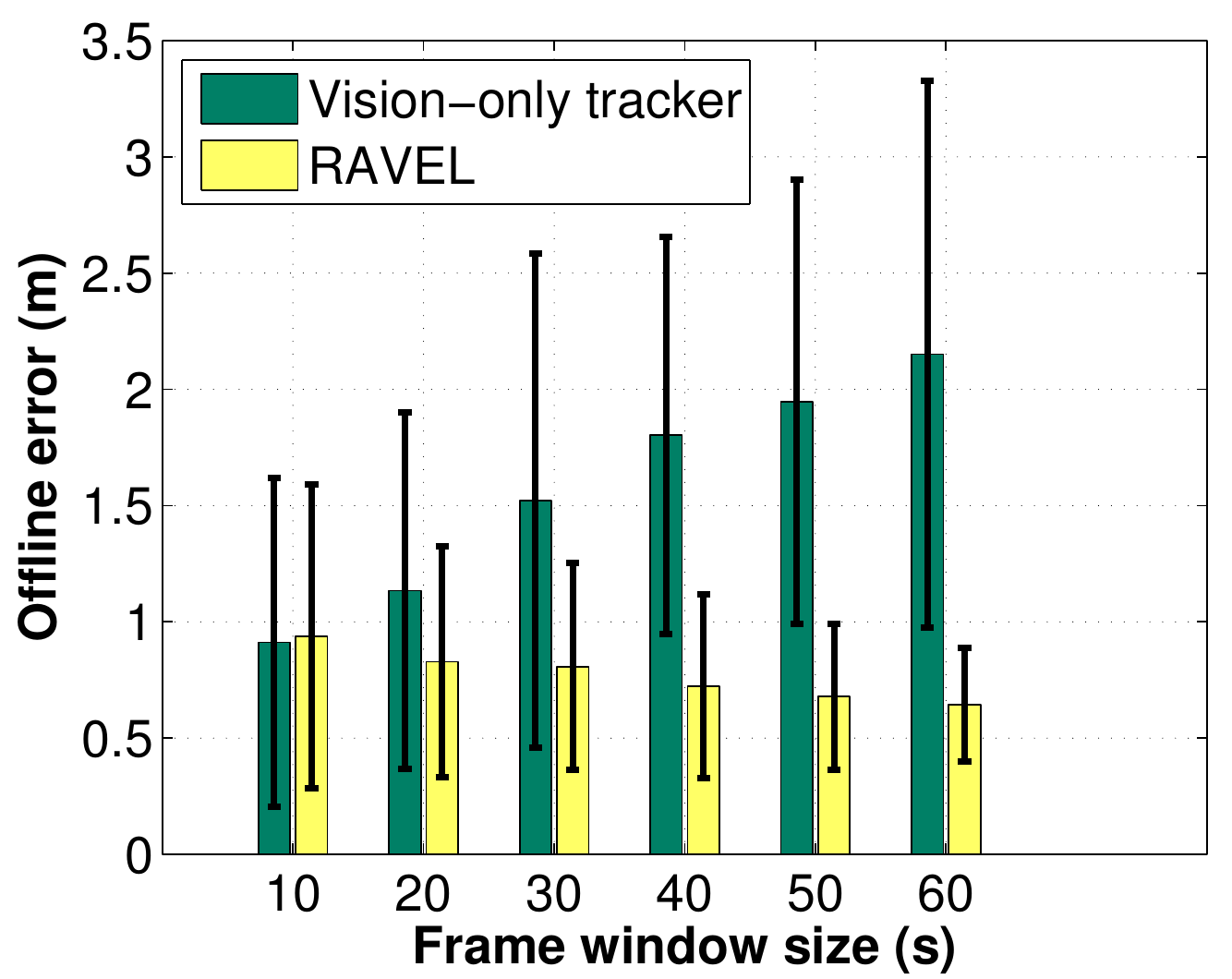}%
\caption{}%
\label{fig:windowSize}%
\end{subfigure}\hfill%
\begin{subfigure}{0.68\columnwidth}
\includegraphics[width=\columnwidth]{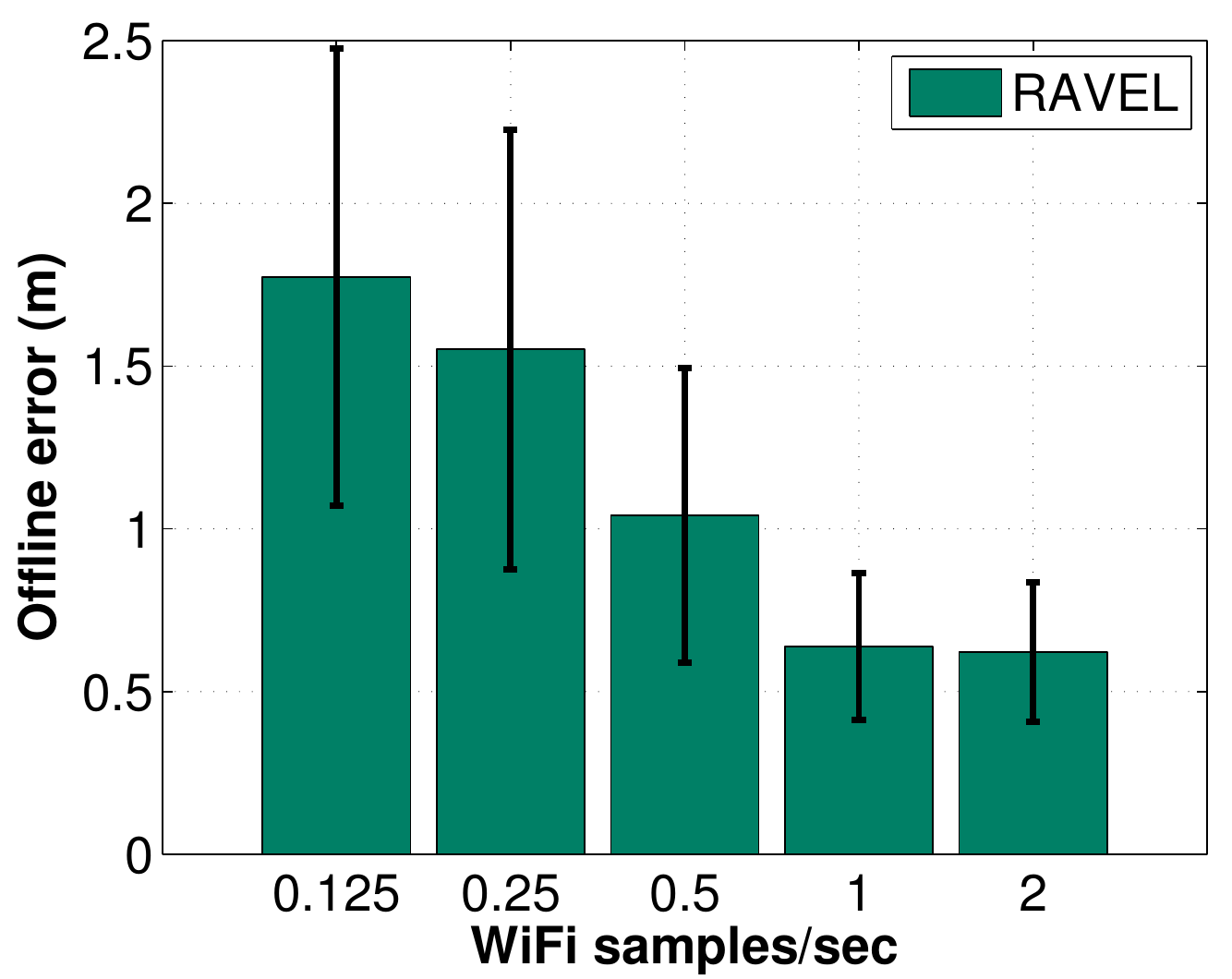}%
\caption{}%
\label{fig:wifiRate}%
\end{subfigure}\hfill%
\caption{(a) Cumulative distribution function of the offline location error. (b) Impact of the frame window size on the offline location error. (c) Impact of the WiFi sampling rate on the offline location error.}
\vspace*{-2mm}
\end{figure*}

\subsection{Results}

%\begin{figure}
%\centering
%\begin{subfigure}{0.5\columnwidth}
%\includegraphics[width=\columnwidth]{wifiModel.pdf}
%\caption{}%
%\label{subfigWifiCont}%
%\end{subfigure}\hfill%
%\begin{subfigure}{0.48\columnwidth}
%\includegraphics[width=\columnwidth]{regress.pdf}
%\caption{}%
%\label{subfigRegress}%
%\end{subfigure}\hfill%
%\caption{(a) Learning the radio model parameters, (b) Ground truth radio model from 4 APs.}
%\label{}
%\end{figure}

\noindent {\bf No a priori knowledge of the radio model:} The first set of experiments is conducted to validate our assumption that we do not require a priori knowledge of the parameters of the radio propagation model. Our proposed algorithm allows us to learn the model from camera and WiFi data, by considering different radio models (from a prior distribution) and choosing the one that maximizes the fitness function described in Eqn.~\eqref{eq:likelihood-function}. This is in contrast to EV-Loc which assumes a priori knowledge of the radio propagation model and obtains it by collecting training data from the environment (WiFi signal strength data at known distances from the access points). We refer to the radio model learnt from our proposed approach as \emph{Learned model} and the model derived from training data as \emph{Fitted model}. 

Our proposed algorithm, RAVEL, assumes a non-environment specific raised cosine prior distribution on the two parameters ($P_0$ and $n$) of the radio propagation model (see Eqn.~\eqref{eq:radio-model}), with a generous variance for each parameter. The prior distributions for $P_0$ and $n$ are based on smartphone-based WiFi specs and a number of studies on typical values of $n$, e.g. \cite{Seidel:IEEEtran:1992}. Fig. (\ref{fig:learnModel}a) shows the received signal strength measurements (600 samples are shown) as we vary the log-distance from four access points and the fitted ground truth radio propagation model. The fitted model ($P_0=-64$ and $n=2.2$) is obtained by averaging the $P_0$ and $n$ values derived from the training data of the four access points.

Fig. (\ref{fig:learnModel}b) shows that the fitness of the best hypothesis (given by Eqn.~\eqref{eq:likelihood-function}) as we vary the parameters of the radio propagation model. Note that the Learned model, derived from the proposed RAVEL algorithm by maximizing the fitness of the best hypothesis, is very close to the \emph{Fitted model} obtained from a wealth of training data (an expensive survey). More importantly this Learned model only assumes a rather broad and uninformative prior, the mode of which (referred to as \emph{Prior model}) is quite far from the Fitted model. Hence, with very little prior knowledge of the radio propagation model, we are able to accurately infer it and avoid the time consuming step of environment-specific training. %However, reliable prior knowledge of the radio model would constraint the search space thus reducing the power requirements of our algorithm making it more suitable for mobile devices and real-time processing.

\noindent {\bf Offline location error:} The second set of experiments focuses on the offline case, and aims to compare the offline location error of RAVEL and the competing Vision-only tracker. We first use default values for the window size (120 frames taken over 60 secs) and for the WiFi sampling rate (2 WiFi signal strength values per sec), and plot the CDF of the offline location error over 160 windows (40 windows for each of the four people carrying a smartphone with WiFi). Fig.~(\ref{fig:cdfOfflineError}) shows that RAVEL, which leverages WiFi information, achieves a median error of 0.56 m and a 90 percentile error of 1 m, significantly outperforming the competing Vision-only tracker, which has a median error of about 1 m and a 90 percentile error of 4.6 m. Note that to give a fair chance to the Vision-only tracker, we initialize it with the correct visual detection corresponding to the person that we are tracking. However, even with a correct start the Vision-only tracker diverges from the correct path due to crossing, splitting and other ambiguities. These ambiguities are typically resolved by RAVEL with the help of WiFi data; illustrative examples of how this happens are provided later in the evaluation. 

We then proceeded to examine the impact of the window size on the offline location error. We kept the WiFi sampling rate to the default value of 2 samples per sec, and varied the window size (10-60 secs corresponding to 20-120 frames). Fig. (\ref{fig:windowSize}) shows that the proposed RAVEL approach improves its accuracy as the window size increases. This is reasonable since different people will most likely walk different paths over long periods of time resulting in unique WiFi sequences which act as powerful discriminative signatures for each person. Thus as the frame window size increases the accuracy tends to increase.
The converse is true for the Vision-only tracker. For short periods of time we can see that it can be quite accurate and can achieve similar performance with RAVEL. However, for long periods of time a lot of ambiguities are introduced especially in crowded environments which degrades the performance of Vision-only tracking. 

The next step was to explore the impact of the WiFi RSS sampling rate on the offline location error of RAVEL with a default window size of 120 frames. As expected, Fig.~(\ref{fig:wifiRate}) shows that the performance of RAVEL improves as we increase the sampling rate. Note that no significant benefits are observed by sampling above 1 WiFi RSS per second, which means that radio sampling does not need to be intensive (i.e. the algorithm could run on battery operated devices) to obtain accurate trajectories. This however may be also due to the fact that people walk relatively slowly in museums; it is possible that different environments may benefit from higher sampling rates.

\begin{figure}
\centering
\begin{subfigure}{0.5\columnwidth}
\includegraphics[width=\columnwidth]{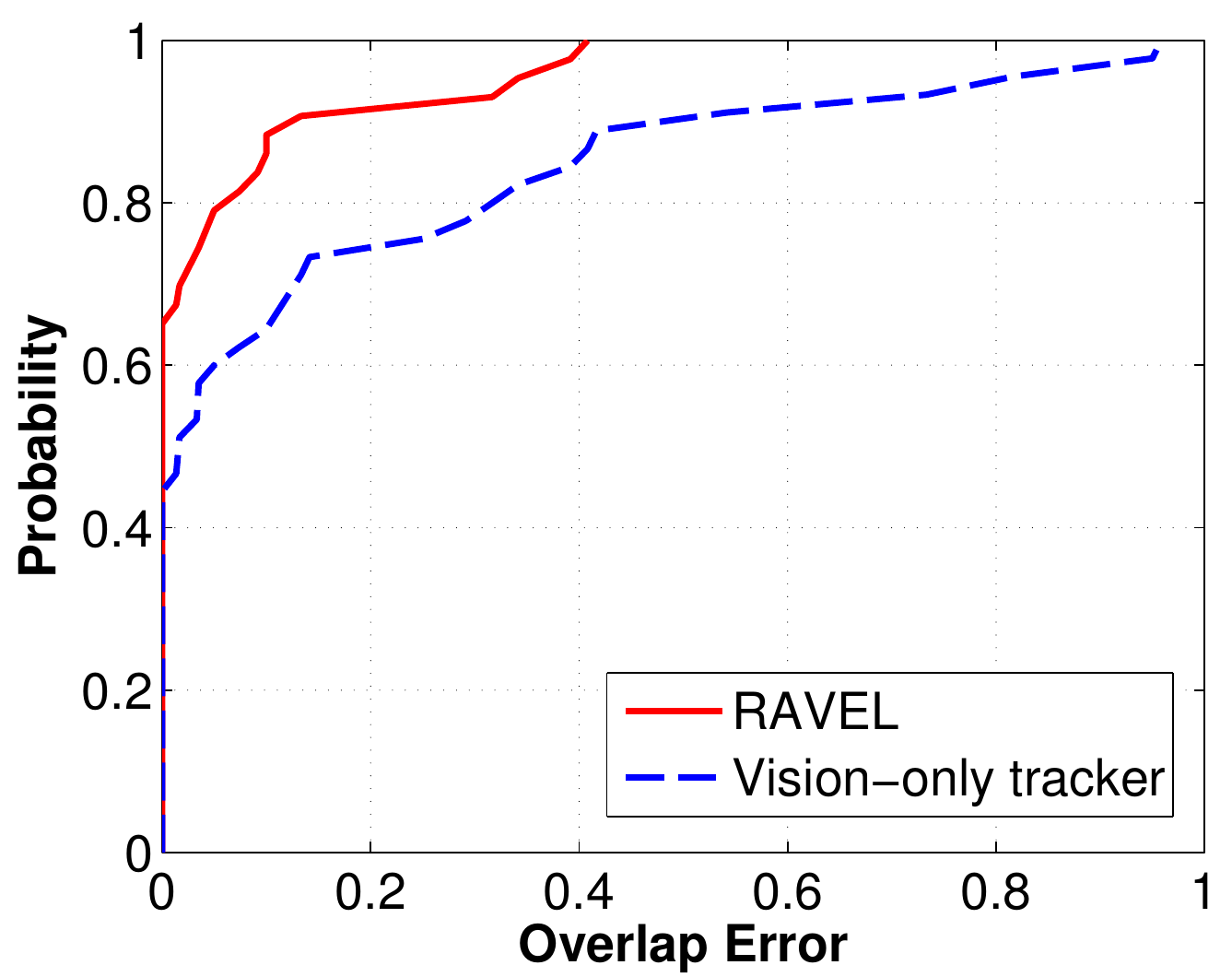}
\caption{}%
\label{fig:cdfOfflineOverlap}%
\end{subfigure}\hfill%
\begin{subfigure}{0.48\columnwidth}
\includegraphics[width=\columnwidth]{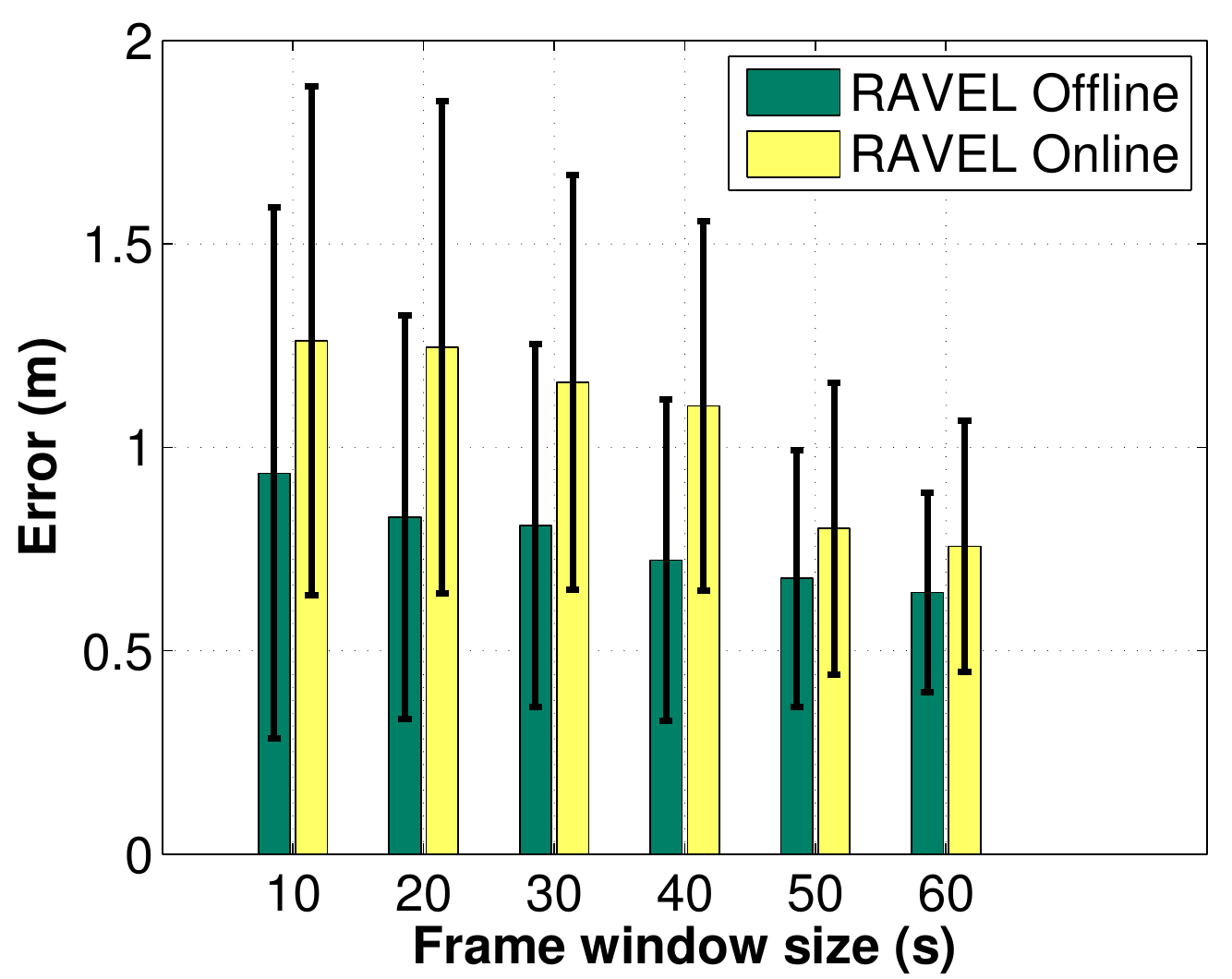}
\caption{}%
\label{fig:onlineError}%
\end{subfigure}\hfill%
\caption{(a) Cumulative distribution function (CDF) of the overlap error. (b) Online location error.}
\label{}
\vspace*{-1 mm}
\end{figure}

\noindent {\bf Offline overlap error:} The third set of experiments aims to compare RAVEL with Vision-only tracker in terms of their ability to correctly determine whether a target is inside or outside the camera's FOV. To do so, we use the \emph{overlap error} metric defined above as the percentage of frames in which the estimated trajectory incorrectly places the user inside or outside the FOV when compared to the ground truth. Fig.~(\ref{fig:cdfOfflineOverlap}) shows the CDF of the overlap error of the two algorithms over 160 windows (40 non-overlapping windows times four people). Observe that up to 70\% of these windows have no overlap error for RAVEL, as opposed to 50\% for the competing approach. The maximum overlap error of RAVEL is 40\% as opposed to 100\% for Vision-only tracker. This shows that RAVEL's superior performance in resolving ambiguities that result from people entering and leaving the FOV and re-appearing into the FOV from different entry points. Illustrative examples of such cases will be examined below.

%\begin{figure}
%\centering
%\includegraphics[width=\columnwidth]{cdfOVERLAP.pdf}
%\caption{Cumulative distribution function (CDF) of the overlap error}%
%\label{fig:cdfOfflineOverlap}
%\end{figure}

\noindent {\bf Online location error:} The last set of experiments focuses on the online case, where we are interested in the most recent location of a target. We still assume that we have access to a historical window of visual and WiFi detections, and we investigate the impact of the window size on the online (most recent) location error. As expected, Fig.~(\ref{fig:onlineError}) shows that the online location error decreases as we increase the window size. This behavior is similar to that of the offline location error (also shown in Fig.~(\ref{fig:windowSize})), which measures the average error across all positions of the window. Notice in Fig.~(\ref{fig:windowSize}) that the online location error is always slightly higher than the offline location error for a given window size, because the last position estimate in a given window has no future detections to benefit from.

%\begin{figure}
%\centering
%\includegraphics[width=\columnwidth]{onlineError.pdf}
%\caption{Online location error}%
%\label{fig:onlineError}
%\end{figure}

\begin{figure*}
\centering
\begin{subfigure}{0.68\columnwidth}
\includegraphics[width=\columnwidth]{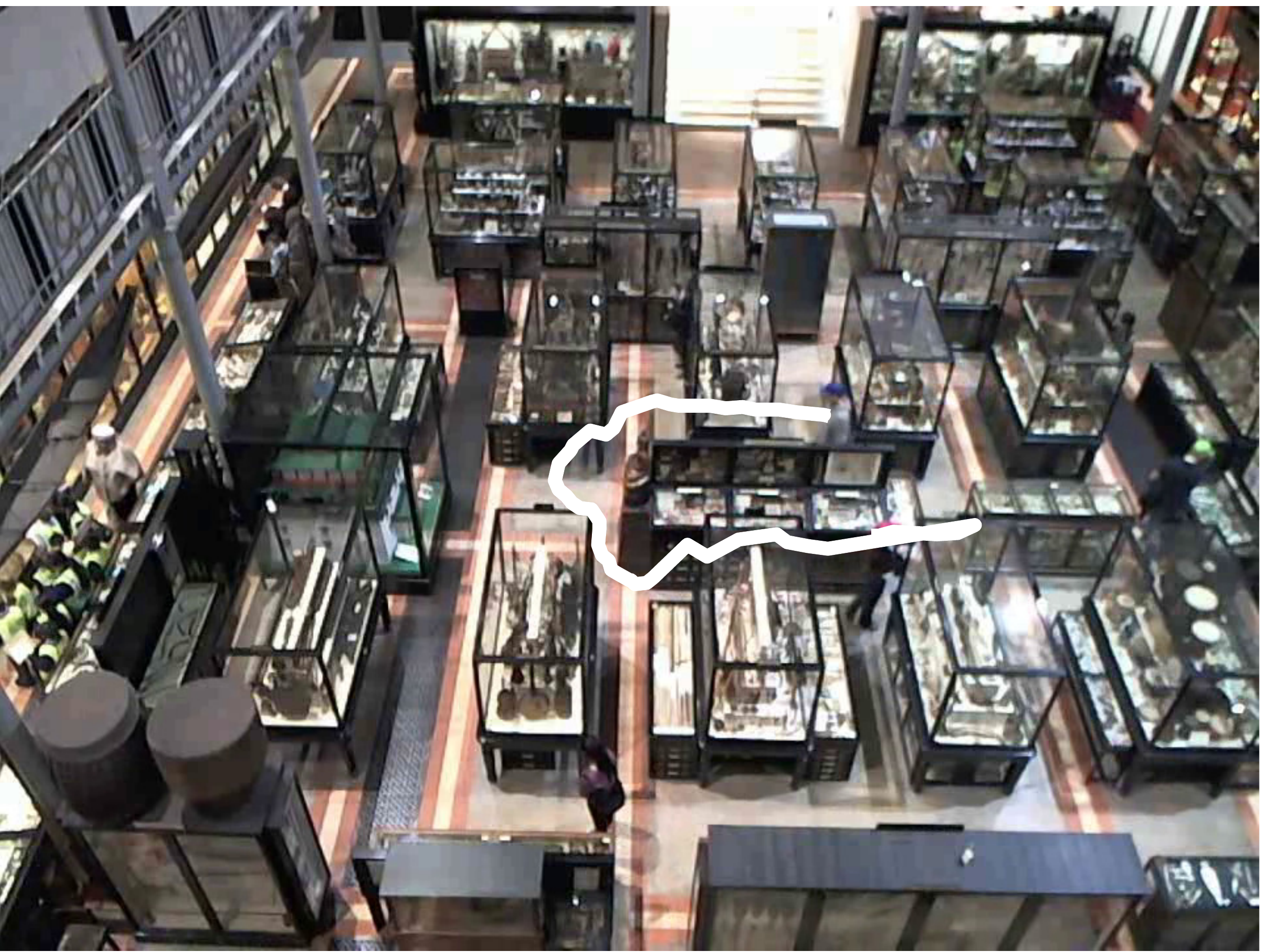}%
\caption{Ground truth}%
\label{subfiga}%
\end{subfigure}\hfill
\begin{subfigure}{0.68\columnwidth}
\includegraphics[width=\columnwidth]{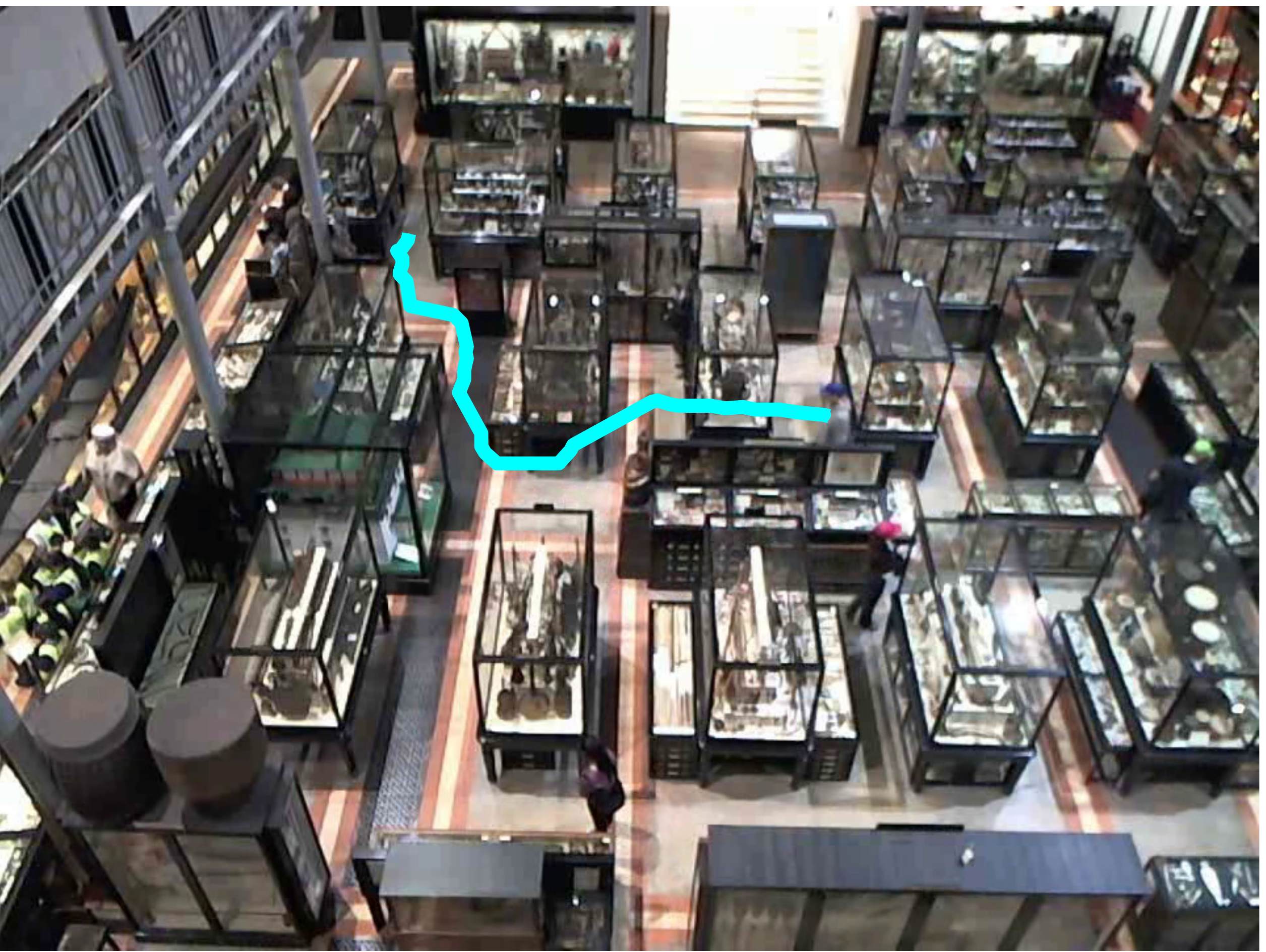}%
\caption{Vision-only tracker}%
\label{subfigb}%
\end{subfigure}\hfill
\begin{subfigure}{0.68\columnwidth}
\includegraphics[width=\columnwidth]{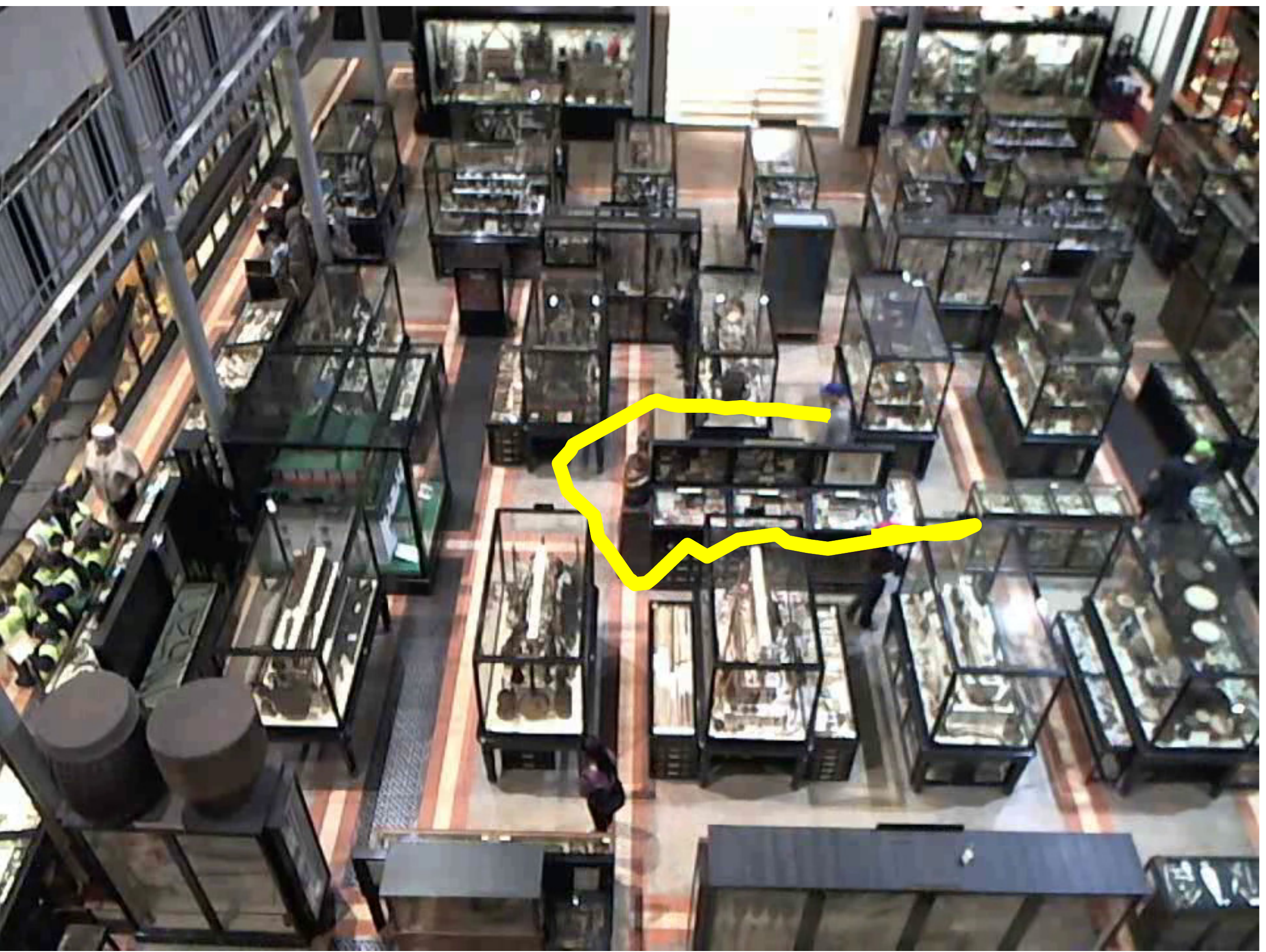}%
\caption{RAVEL}%
\label{subfigc}%
\end{subfigure}\hfill

\begin{subfigure}{0.68\columnwidth}
\includegraphics[width=\columnwidth]{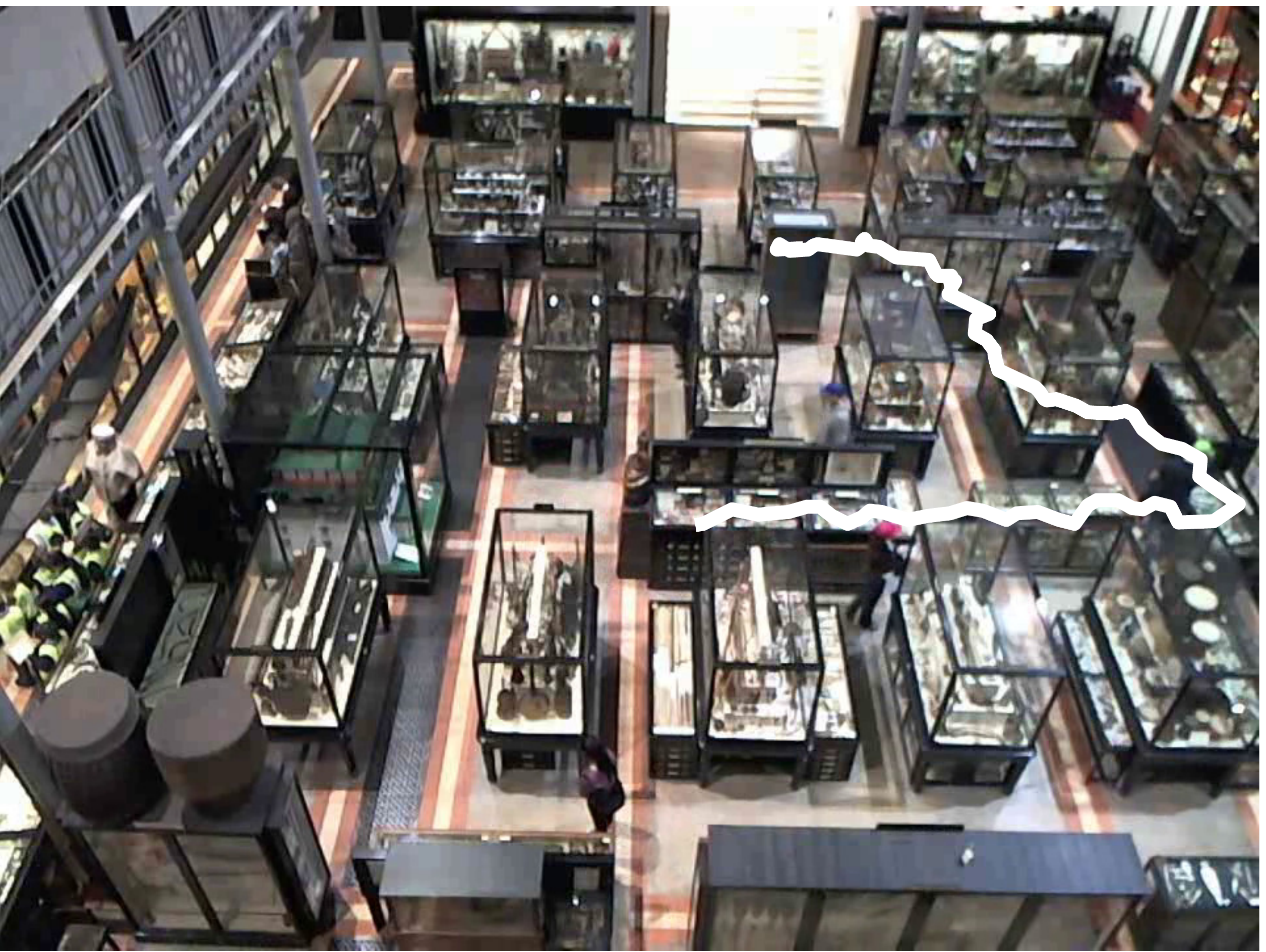}%
\caption{Ground truth}%
\label{subfigd}%
\end{subfigure}\hfill
\begin{subfigure}{0.68\columnwidth}
\includegraphics[width=\columnwidth]{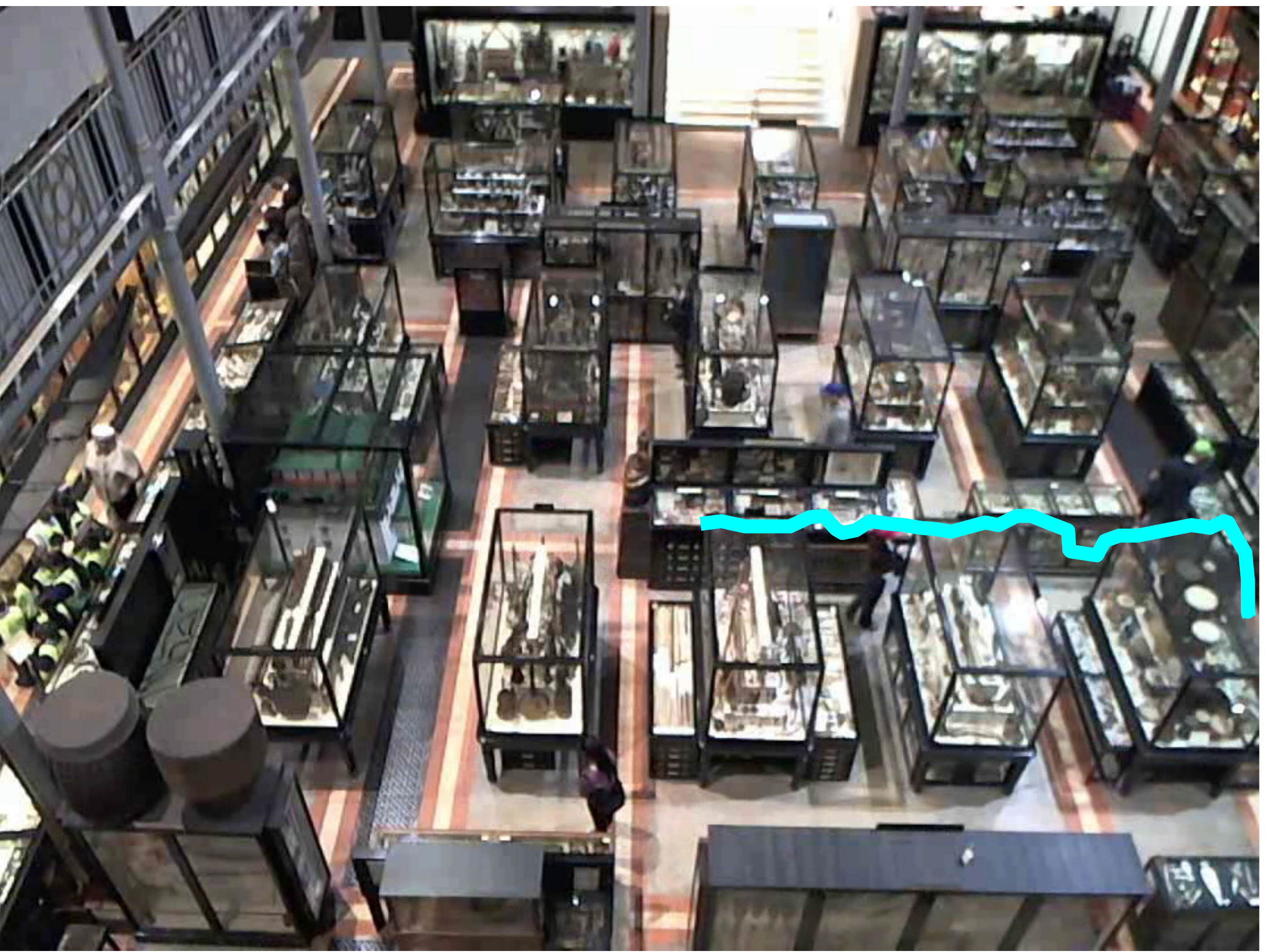}%
\caption{Vision-only tracker}%
\label{subfige}%
\end{subfigure}\hfill
\begin{subfigure}{0.68\columnwidth}
\includegraphics[width=\columnwidth]{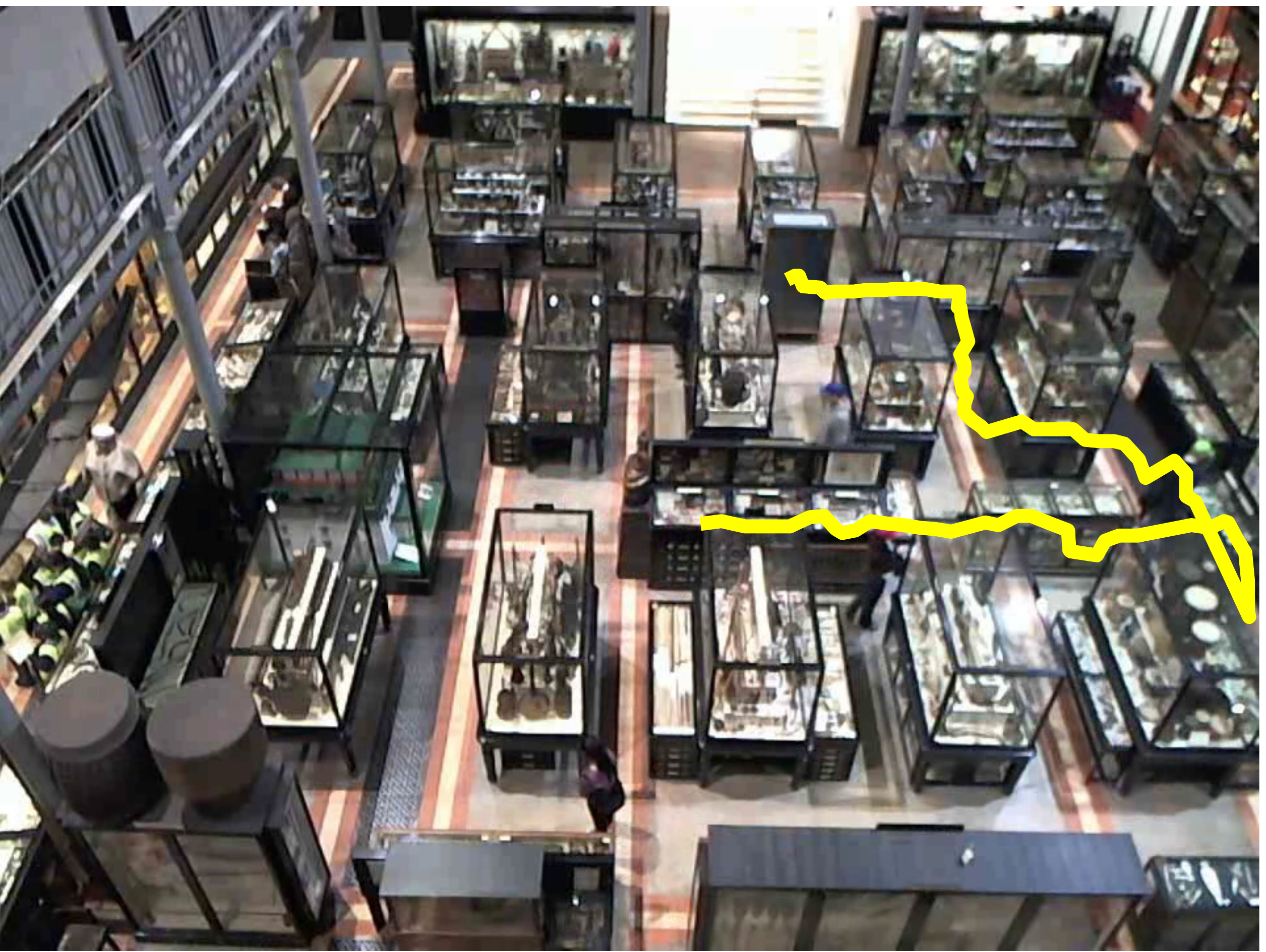}%
\caption{RAVEL}%
\label{subfigf}%
\end{subfigure}\hfill

\begin{subfigure}{0.68\columnwidth}
\includegraphics[width=\columnwidth]{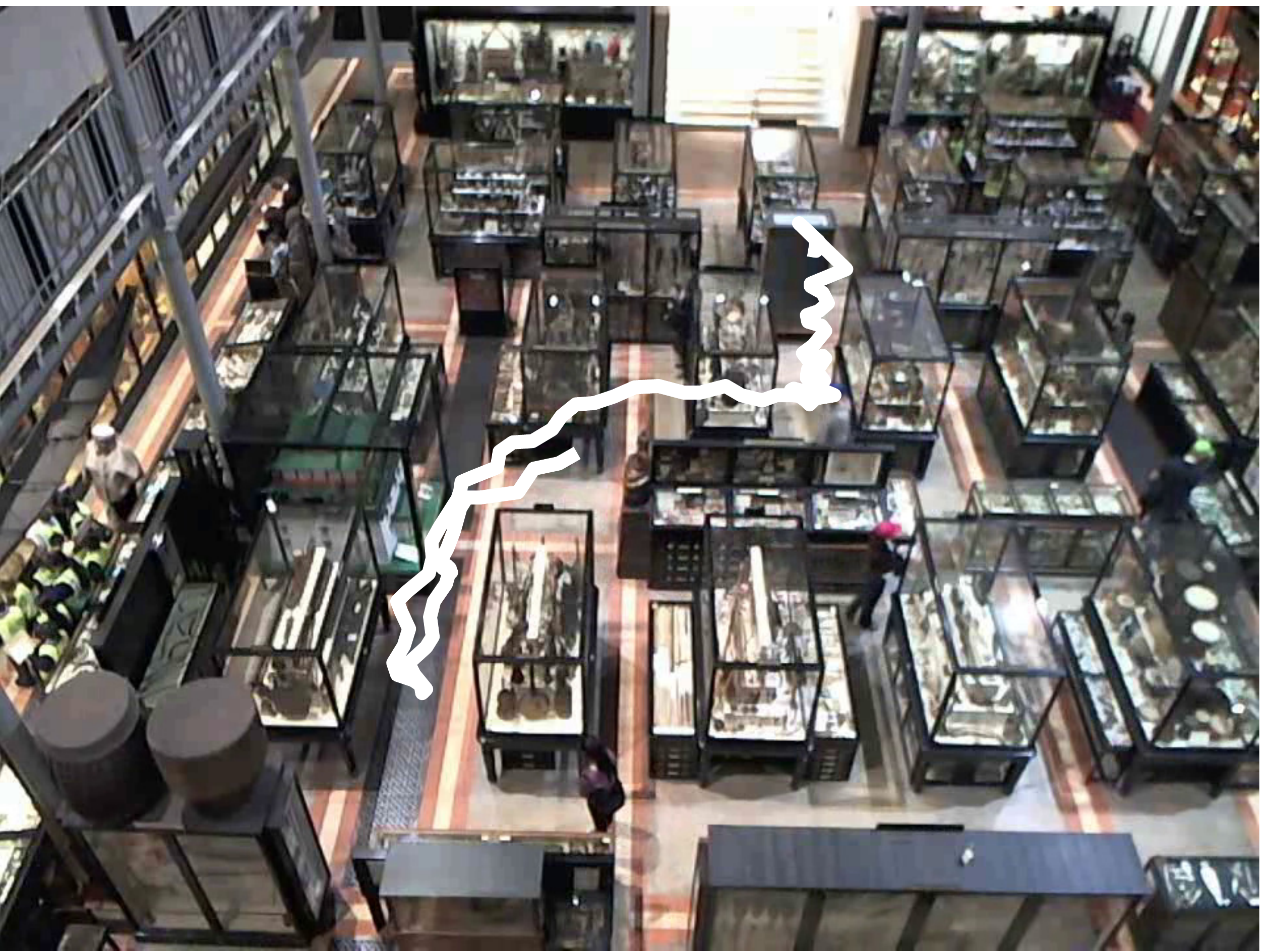}%
\caption{Ground truth}%
\label{subfigg}%
\end{subfigure}\hfill
\begin{subfigure}{0.68\columnwidth}
\includegraphics[width=\columnwidth]{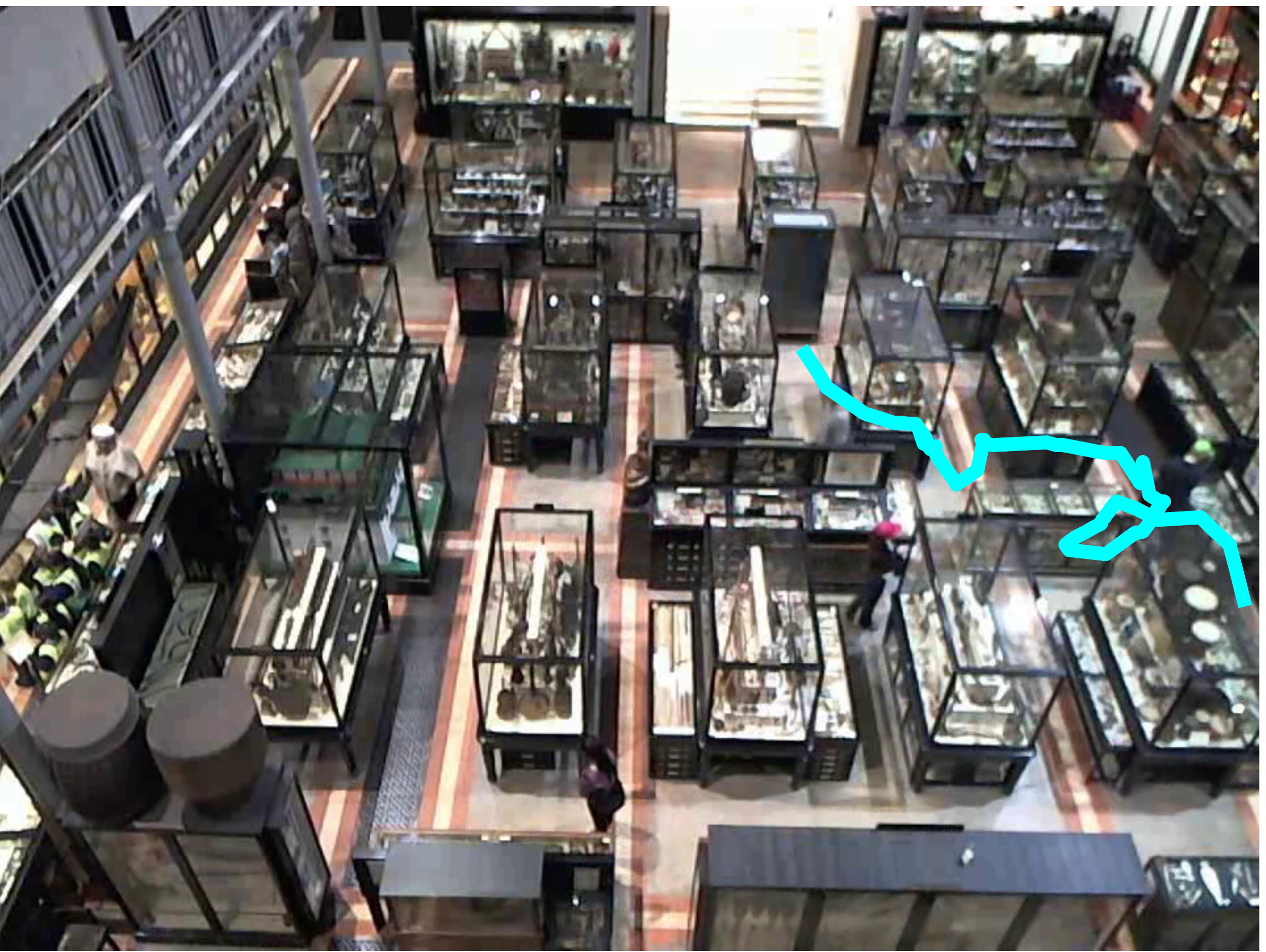}%
\caption{Vision-only tracker}%
\label{subfigh}%
\end{subfigure}\hfill
\begin{subfigure}{0.68\columnwidth}
\includegraphics[width=\columnwidth]{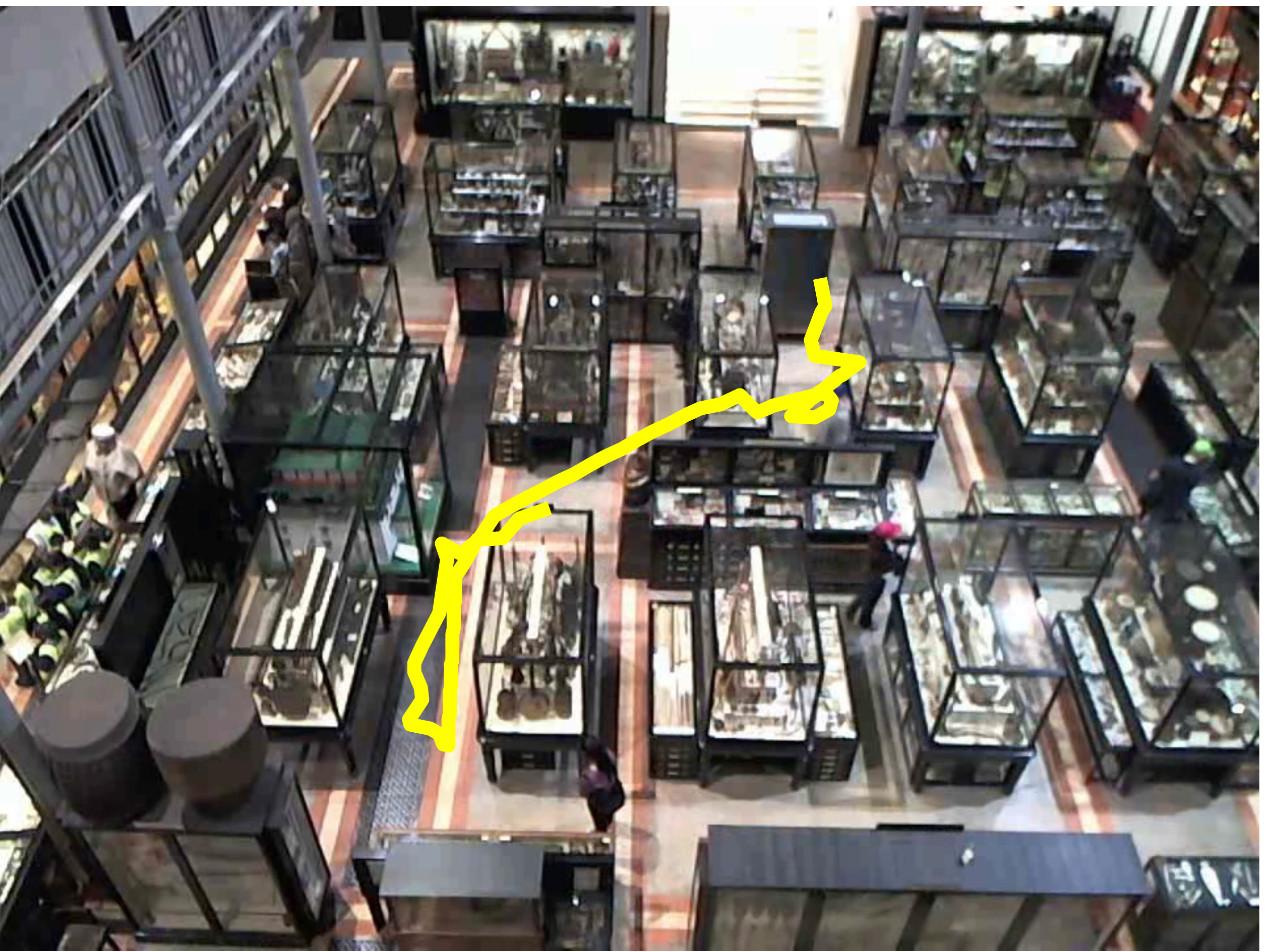}%
\caption{RAVEL}%
\label{subfigi}%
\end{subfigure}\hfill

\caption{Illustrative examples showing the performance of the proposed and competing algorithms.}
\label{cases}
\vspace*{-2mm}
\end{figure*}

\noindent {\bf Illustrative examples:} We close our evaluation by showing in Fig. (\ref{cases}) three illustrative examples showing the performance of the proposed and competing algorithms: In the first example (Fig.~(\ref{subfiga})), the person stops at particular museum exhibit, and stays there for a long period of time. Then she starts walking again, and performs a U turn. While the person is looking at the exhibit without moving, no camera detections are generated. Once she starts moving again it becomes difficult to distinguish in which direction she actually moves using camera detections due to ambiguities with other people's trajectories. One hypothesis is that her path merged with a second person's path moving north as illustrated in Fig.~(\ref{subfigb}); another hypothesis is that she makes a U turn as in Fig.~(\ref{subfigc}). By taking into account her radio data, RAVEL can identify the correct trajectory. Another interesting case where WiFi measurements are beneficial is illustrated in Fig.~(\ref{subfigd}). In this case the person leaves the camera's FOV and after some time he re-enters. Fig.~(\ref{subfigf}) shows that RAVEL can re-establish the identity of the person when he re-enters the FOV and provide the correct trajectory. Finally, Fig.~(\ref{subfigg}) illustrates the situation of two targets splitting paths which is quite common in crowded environments. More specifically, two people start walking together (i.e. one detection contains two targets) moving south when at some point they split following different paths. In this case the Vision-only tracker has a 50\% chance of following the wrong path, whereas RAVEL can use radio data to select the correct path. Although RAVEL allows us to resolve a number of common ambiguities, it is by no means a perfect tracking system; we have observed certain cases in which ambiguities are not correctly resolved either due to a very limited window size, or to a highly crowded and complex scene. In addition, if there are regions where either WiFi or vision are unavailable, in its current implementation no positions are returned. In the future this limitation could be solved by incorporating other modalities such as inertial sensing. Our qualitative and quantitative results though suggest that RAVEL is significantly more accurate and robust than existing Vision-only tracking systems, requiring only a negligible overhead to extend their implementation.

%\flushcolsend
%%% Local Variables: 
%%% mode: latex
%%% TeX-master: "main"
%%% End: 

%% file: background.tex
%%%%%%%%%%%%%%%%%%%%%%%%%%%%%%%%%%%%%%%%%%
% Background

\section{Related Work}
\label{sec:background}
Extensive research has been done on localization techniques, and a recent survey can be found in~\cite{Mautz:IPS-Book:2012}. Here we provide a brief survey focusing only on state of the art techniques that combine vision with other sensing modalities, such as radio and inertial sensors. Mandeljc et al.~\cite{Mandeljc:ICDSC:2011} recently proposed a fusion algorithm that incorporates radio data into a camera-based probabilistic occupancy map (POM) framework. As in the original POM framework~\cite{Fleuret:PAMI:2008}, humans are represented as simple rectangles, and detections are generated using a background subtraction technique. The work in~\cite{Mandeljc:Sensors:2012} generalizes POM to work with arbitrary sensor modalities, and provides an illustrative example with Ubisense's ultra-wideband radio sensors. However, unlike our work, radio measurements are only applied to the sub-problem of estimating a ground plane occupancy model (that gives the probability of a cell being occupied by any of the humans given sensor data); they are not directly used for identification, i.e. to estimate the probability of a particular person being located in a cell given that the cell is occupied. The latter problem is addressed in~\cite{Fleuret:PAMI:2008,Mandeljc:ICDSC:2011} using the color distributions of camera detections, without the help of radio. Color distributions and other appearance features are of little use in our context due to the challenging light conditions and the downward facing camera configuration. Hence, these works do not exploit the full power of radio for disambiguating human trajectories. More recent work~\cite{Mandeljc:Sensors:2012} added a second fusion stage, where anonymous detections are augmented with identity information from radio tags. The cost of mapping an anonymous detection to a radio-based identified detection is evaluated based on the Euclidean distance between the two, and the optimal assignment between two sets of detections in a frame is evaluated using the Hungarian method~\cite{Kuhn:NRLQ:1955}. Although this approach~\cite{Mandeljc:Sensors:2012} exploits radio-based detections for identification, it does so on a frame-by-frame basis, and does not jointly address the problem of identification and tracking. More importantly, it assumes knowledge of multiple people's radio data (instead of one person's in our work), and thus addresses a different problem. 

Other techniques~\cite{Jung:INFOCOM:2010,Teixeira:PETRA:2009} fuse camera and inertial measurements. More specifically, the authors in~\cite{Jung:INFOCOM:2010,Teixeira:PETRA:2009} use cameras and accelerometers to identify people based on their velocity. The correlation coefficient is being used to calculate the similarity between the velocity of each user estimated from camera and accelerometer measurements. This work is orthogonal to the proposed RAVEL system, and the two papers could be combined in the future to exploit all three sensing modalities (vision, radio and inertial sensors). 

Finally the recent EV-Loc~\cite{Zhang:MobiHoc:2012} system demonstrated how to perform localization by combining electronic and visual signals. Given a number of people equipped with mobile phones, the algorithm estimates the position of each person using both WiFi signal strength and visual tracking. Then an optimization problem must be solved in order to find the best mapping between electronic and visual signals. Finally, the position of each person is calculated as the weighted average of the corresponding electronic and visual estimates. As we already mentioned in the previous sections EV-Loc assumes that the vision system provides clear visual trajectories (i.e. the data association problem is solved) and thus this system concentrates on finding the association between WiFi traces and visual traces given a known radio propagation model. Similar to~\cite{Mandeljc:Sensors:2012}, it assumes knowledge of multiple people's radio data and thus addresses a slightly different problem than the one tackled by RAVEL.

%%% Local Variables: 
%%% mode: latex
%%% TeX-master: "main"
%%% End: 

%% file: conclusion.tex
%%%%%%%%%%%%%%%%%%%%%%%%%%%%%%%%%%%%%%%%%%
% Conclusion

\section{Conclusion}
\label{sec:conclusion}

We have introduced RAVEL, a new positioning system that integrates anonymous visual tracking with radio measurements to provide accurate tracking information. Although noisy radio measurements are inadequate for positioning on their own, we demonstrated how they can greatly enhance the performance of visual tracking by handling challenging cases like occlusion, trajectory splitting and entry/exit from the camera field-of-view. The additional advantage of our approach is that it automatically learns the radio propagation model, thus not requiring site-specific calibration. We believe that this work has the potential to provide highly accurate positioning at low cost in challenging indoor environments. 

%We are currently working on incorporating our tracking algorithm in IP cameras.

%We propose a new positioning system that integrates WiFi information with visual tracking. The novelty of our approach is that it leverages WiFi measurements to improve the performance of visual tracking, and offer accurate localization and identification at the same time. 

%%% Local Variables: 
%%% mode: latex
%%% TeX-master: "main"
%%% End: 